\RequirePackage{fix-cm}
\documentclass[twocolumn]{svjour3}          
\smartqed  
\usepackage{graphicx}
\usepackage{natbib}
\usepackage{multirow}
\usepackage{amsmath}
\usepackage{amssymb}
\usepackage{bbm}
\usepackage{bm}
\usepackage{booktabs}
\usepackage{array} 
\usepackage{blindtext}
\usepackage{comment}
\usepackage[breaklinks=true]{hyperref}

\usepackage{pifont}

\usepackage[dvipsnames]{xcolor}
\usepackage{color, colortbl}
\definecolor{citecolor}{HTML}{0071bc}
\definecolor{tabhighlight}{HTML}{e5e5e5}
\newcommand{\myparagraph}[1]{\vspace{0.1em}\noindent\textbf{#1}}
\makeatletter
\renewcommand\paragraph{
  \@startsection{paragraph} 
  {4} 
  {\z@} 
  {.5em \@plus1ex \@minus.2ex} 
  {-.5em} 
  {\normalfont\normalsize\bfseries} 
}
\makeatother
\begin{document}
\sloppy

\title{InterGen: Diffusion-based Multi-human Motion Generation under Complex Interactions 
}

\author{Han Liang$^{1}$ \and
        Wenqian Zhang$^{1}$ \and
        Wenxuan Li$^{1}$ \and
        Jingyi Yu$^1$ \and
        Lan Xu$^1$ 
}



\institute{Han Liang \at
              \email{lianghan@shanghaitech.edu.cn}
           \and
           Wenqian Zhang \at
              \email{zhangwq2022@shanghaitech.edu.cn}
           \and
           Wenxuan Li \at
              \email{liwx2@shanghaitech.edu.cn}
           \and
            Jingyi Yu\at
              \email{yujingyi@shanghaitech.edu.cn}
           \and
           Lan Xu \at
              \email{xulan1@shanghaitech.edu.cn}
           \\
           $^1$ ShanghaiTech University, Shanghai, China
           \\
}
\date{Received: date / Accepted: date}
\maketitle
\begin{abstract} \label{abstract}
We have recently seen tremendous progress in diffusion advances for generating realistic human motions. Yet, they largely disregard the multi-human interactions. 
In this paper, we present InterGen, an effective diffusion-based approach that enables layman users to customize high-quality two-person interaction motions, with only text guidance.
We first contribute a multimodal dataset, named InterHuman. It consists of about 107M frames for diverse two-person interactions, with accurate skeletal motions and 23,337 natural language descriptions.
For the algorithm side, we carefully tailor the motion diffusion model to our two-person interaction setting. 
To handle the symmetry of human identities during interactions, we propose two cooperative transformer-based denoisers that explicitly share weights, with a mutual attention mechanism to further connect the two denoising processes.
Then, we propose a novel representation for motion input in our interaction diffusion model, which explicitly formulates the global relations between the two performers in the world frame.
We further introduce two novel regularization terms to encode spatial relations, equipped with a corresponding damping scheme during the training of our interaction diffusion model. 
Extensive experiments validate the effectiveness of InterGen\footnote{\url{https://tr3e.github.io/intergen-page/}}. Notably, it can generate more diverse and compelling two-person motions than previous methods and enables various downstream applications for human interactions.

\keywords{Motion synthesis  \and Multimodal generation \and  Diffusion model  \and  Text-driven generation}

\end{abstract}

\section{Introduction} \label{introduction}

Digital human motions should reflect how we humans interact and communicate with each other, in order to depict the diverse cultures and societies that make up our physical world. A successful motion creation tool will hence allow users to customize realistic human motions under interactions. The produced motions also need to match specific themes, e.g., from as complicated as the movie script, or as simple as textual descriptions by novice users. Such human motion generation serves as a core computer vision problem, with various applications in VR/AR, games, or films.

Recent years have witnessed impressive progress in human motion generation under various user-specified conditioning, such as action categories~\citep{guo2020action2motion,petrovich2021action}, music pieces~\citep{li2022danceformer,li2021ai}, speeches~\citep{habibie2022motion,ao2022rhythmic}, or natural text prompts~\citep{petrovich2022temos,tevet2022human}.
The key idea is to learn a conditional generative model for the complex multimodal distribution of human motions, equipped with powerful neural techniques, from variational autoencoders (VAEs)~\citep{kingma2013auto}, generative adversarial networks (GANs)~\citep{goodfellow2020generative}, normalization flows~\citep{rezende2015variational}, to the latest Diffusion Models~\citep{ho2020denoising,song2020score}.
Only recently, the elaborate large-scale language models~\citep{devlin2018bert,brown2020language} and diffusion methods~\citep{ho2020denoising,you2020graph} have quickly found their way into this field, due to their natural and convenient input controls and the strong ability to model complex distributions.
These prompt-guided motion diffusion methods~\citep{yuan2022physdiff,zhang2022motiondiffuse,tevet2022human,chen2022executing} significantly democratize the accessible and high-quality generation of human motions for novices.

However, most of the above motion diffusion models are tailored for single-person setting, and hence overlook one essential aspect of human motions – the rich human-to-human interactions. The challenges are manifold. First, existing datasets~\cite{punnakkal2021babel,liu2019ntu} fail to simultaneously provide accurate captured results and natural prompt labels for diverse human interactions. The former usually relies on expensive dome-like capturing devices while the latter requires tedious and costly manual labeling. As a result, existing methods focus on generating the kinematic structure of a single human body, without exploring the diverse and complex spatial relationships between various human identities during interactions. The recent concurrent work~\cite{shafir2023human} fine-tunes the single-person motion generator MDM~\cite{tevet2022human} into two-person scenario. Yet, it still suffers from unnatural interactions, inherently due to the limited interaction patterns in the single-person training datasets. In a nutshell, the lack of both multimodal datasets and corresponding explicit modeling schemes constitute barriers in two-person text-guided motion generation.

In this paper, we tackle the above challenges and present \emph{InterGen} -- an effective diffusion-based approach that enables layman users to customize high-quality two-person interaction motions with only text guidance. 
Specifically, we first contribute a novel multimodal human motion dataset, named \emph{InterHuman}, covering a wide range of two-person interactions, from daily ones like hugging, to professional motions, i.e., boxing or Latin dancing. Speficically, we record dense video sequences with 76 RGB cameras, resulting in about 107 million video frames for 7779 motion sequences that last for 6.56 hours. We then recover the ground-truth human skeletal motions under interactions from such rich RGB image modalities using off-the-shelf motion capture approach~\citep{challencap}. Besides, we provide natural language labels for the captured motions, with 23,337 unique descriptions composed of 5656 distinct words. Note that our InterHuman dataset is the first of its kind to open up the research direction for prompt-guided two-person motion generation under interaction setting. Its multi-modality also brings huge potential for future direction like multi-modal human interaction and behavior analysis.

\begin{figure}[tbp] 
	\centering  
	\includegraphics[width=1.0\linewidth]{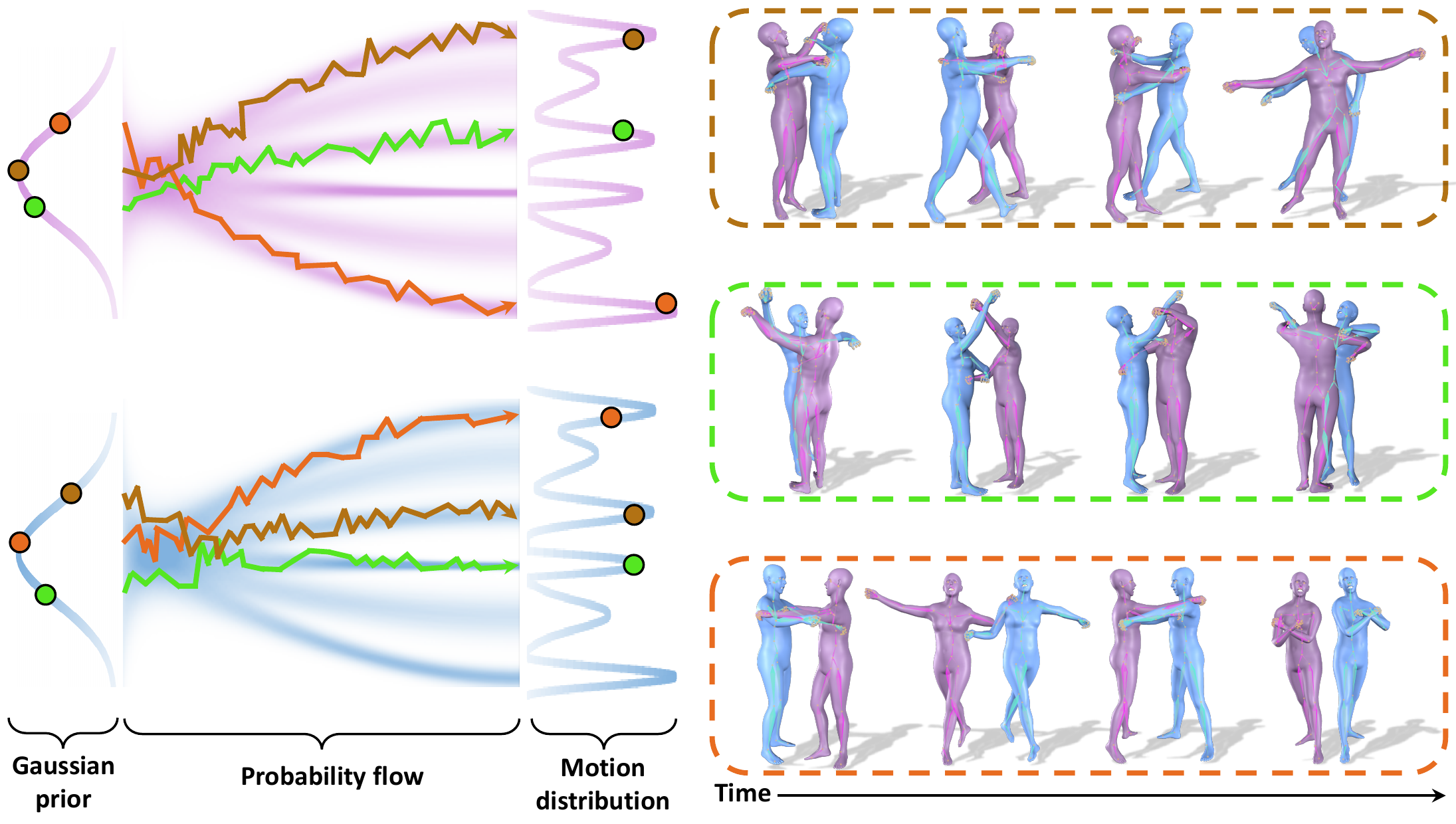} 
	\caption{{InterGen is capable of generating high-quality and diverse motions under complex interactions. It models the two-person symmetry with cooperative diffusion denoisers sharing the same motion manifold.}
} 
 \label{Fig.1} 
\end{figure}

Based on our InterHuman dataset, the key idea of our InterGen approach is to carefully bridge the general diffusion pipeline~\citep{song2020score} into the human motion domain under two-person interactions.
Specifically, we observe the symmetrical fact that exchanging the identities of performers during interactions does not change the semantics of motions. Thus, in our interaction diffusion model, we introduce two cooperative transformer-style denoisers to correspondingly generate the motions of two performers. These denoisers explicitly share weights, with the aid of a novel mutual attention mechanism to further connects the two denoising processes at different feature levels. Such design encourages the two denoisers to perform the same operations and yield the same motion capacity, as illustrated as Fig.~\ref{Fig.1}. Thus, it effectively avoids severe mode collapse when generating interaction motions, e.g., one person can dance professionally while the other cannot.

We further observe the widely adopted canonical representation for single-person motion~\citep{guo2022generating,tevet2022human,zhang2022motiondiffuse} discards the precise spatial relations in our interaction scenarios. Yet, naively adding the relative translation and rotation into the representation will lead to motion drifting during generation. We hence propose a non-canonical motion representation for our interaction diffusion model, where the relations between two people are explicitly encoded by global positions in the same world frame, facilitating the networks to learn relative relations.
Besides, to generate more realistic two-person motions, we introduce two novel regularization terms to model the spatial relations during human-to-human interactions, including a masked joint distance map (DM) loss and relative orientation (RO) loss. The former DM loss encodes the spatial interference between two people with implicit physical constraints, while the latter RO loss encodes the orientation information since we humans pay more attention to our frontal orientation while interacting. We further adapt a damping scheme for these two losses during training, especially when the sampled timestamp of the diffusion process reaches specific thresholds, achieving a more diverse generation. 
Finally, we perform extensive experiments to demonstrate that our approach can generate more compelling two-person motions than previous methods, and showcase its various downstream applications for human interactions, i.e., trajectory control, interactive motion inbetweening, and person-to-person generation.

To summarize, our main contributions include: 
\begin{itemize} 
\setlength\itemsep{0em}
    \item We contribute a new human interaction dataset with rich text/motion modalities, and present a novel diffusion-based approach to generate realistic two-person motions from only prompt inputs. 

    \item In our interaction diffusion model, we introduce cooperative denoisers with novel weights-sharing and a mutual attention mechanism to significantly improve the generation quality.
	
    \item We propose an effective motion representation, as well as two additional regularization losses with a damping schedule to model the complex spatial relation under human-to-human interactions.

\end{itemize}

\section{Related Work}\label{related work}

\begin{figure}[tbp] 

	\centering  
	\includegraphics[width=1.0\linewidth]{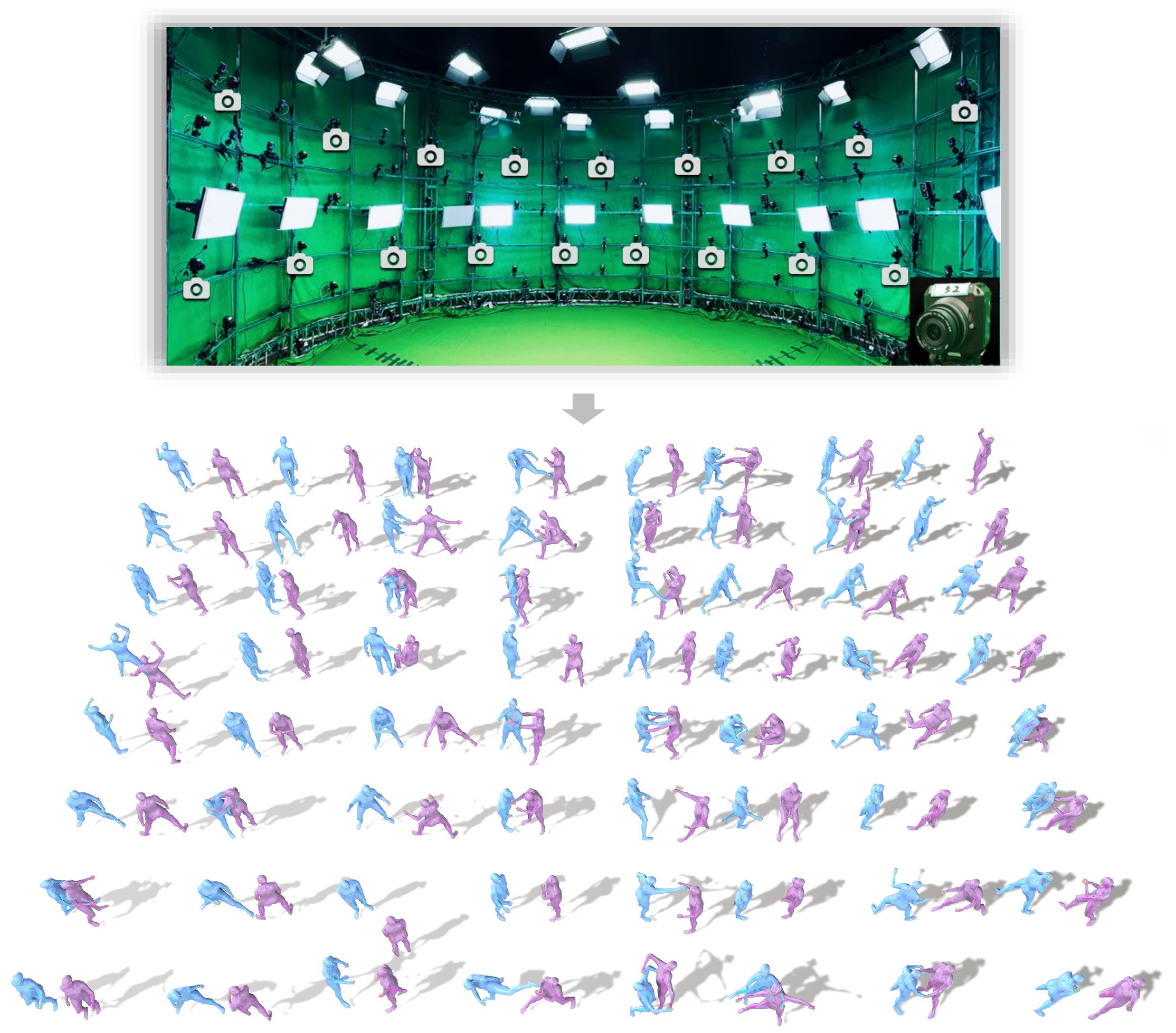}
	\caption{Our motion capture studio (top) and our collected InterHuman dataset illustration (bottom). The system comprises 76 calibrated multi-view cameras. InterHuman covers a wide range of two-
person interactions.} 
 \label{Fig.2} 
\end{figure} 

\subsection{Human Motion Generation}\label{human motion generation}

The field of Human Motion Generation is greatly facilitated by the integration of extensive multimodal data inputs, including text~\citep{petrovich2022temos,tevet2022human,tevet2022motionclip,yuan2022physdiff,chen2022executing,shafir2023human,guo2022generating,kim2022flame}, action ~\citep{guo2020action2motion,petrovich2021action}, incomplete motion sequences~\citep{duan2021single,harvey2020robust}, control signals~\citep{starke2022deepphase,peng2021amp,starke2019neural}, music~\citep{li2021ai,li2022danceformer,lee2019dancing}, speech~\citep{habibie2022motion,ao2022rhythmic}, scene~\citep{wang2021scene,wang2022humanise} and images~\citep{rempe2021humor,chen2022learning}. 

Currently, there exists a variety of works focused on action label-based human motion generation~\citep{guo2020action2motion,petrovich2021action,song2022actformer}. As the emergence of large language models (LLMs)~\citep{brown2020language,touvron2023llama} paves the way for LLM-based multimodal~\citep{openai2023gpt4} models, the potential of text-based multimodal generation becomes increasingly apparent. However, action-based human motion generation is fundamentally a class-based generation approach, which can not enable high-level natural language and multimodal control. Consequently, it is crucial to explore human motion generation based on natural language input. Early approaches such as ~\cite{ahn2018text2action} employ a sequence-to-sequence model to generate upper body motion. Subsequently, ~\cite{ahuja2019language2pose}  and ~\cite{ghosh2021synthesis} concentrate on developing a unified language and pose representation to enable autoregressive synthesis of human movements. Additionally, ~\cite{petrovich2022temos} and ~\cite{guo2022generating} both adopt Variational Autoencoder (VAE)~\citep{kingma2013auto} based architectures for motion generation. 
{\cite{athanasiou2022teach} further condition the VAE on past frames and autoregressively generate an arbitrary sequence of motions given the respective action descriptions. }
Multimodal pre-trained models~\citep{radford2021learning} facilitate more seamless integration between textual and motion spaces~\citep{tevet2022motionclip}.

\begin{figure*}[tbp] 
	\centering  
	\includegraphics[width=1.0\linewidth]{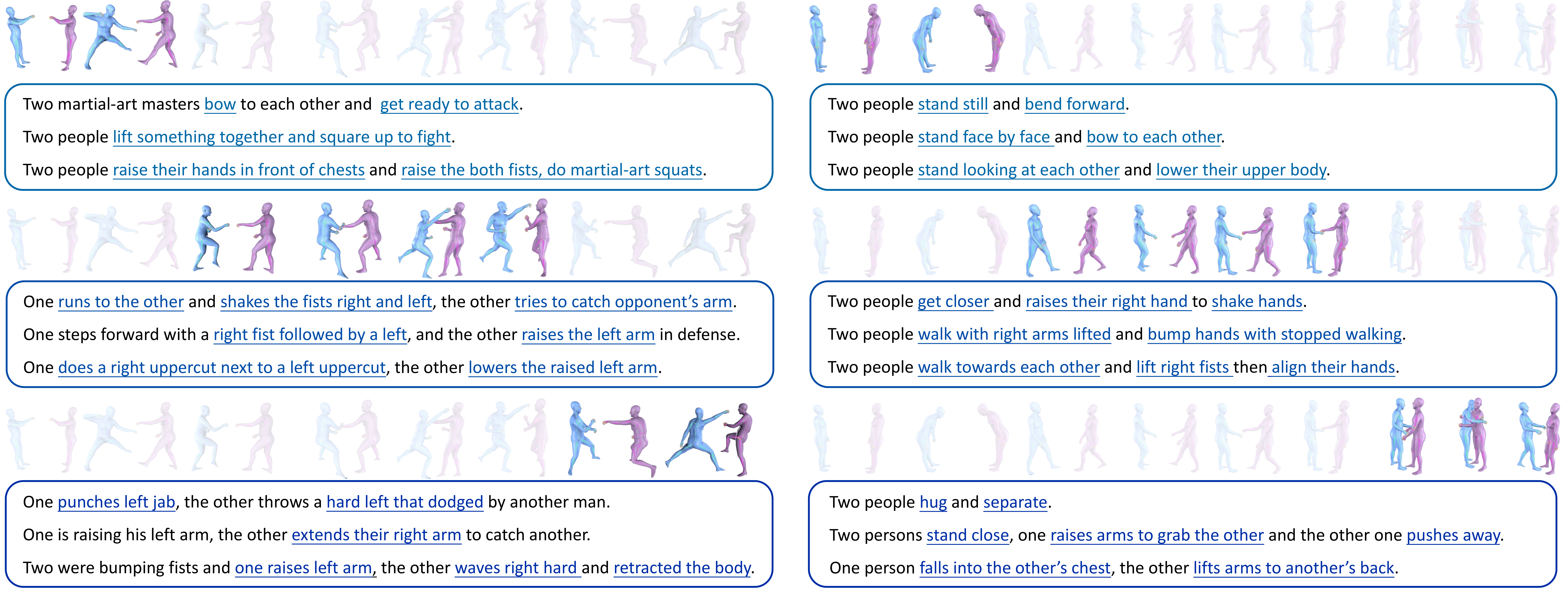} 
	\caption{InterHuman dataset consists of diverse human professional and daily interactions with diverse natural language annotations from different annotators. The figure showcases two examples of our dataset, martial arts, and social manners, with thorough descriptions from different perspectives. } 
    \label{Fig.3} 
\end{figure*}

Recent advancements in diffusion models~\citep{ho2020denoising,song2020score} have significantly propelled text-driven motion generation.
~\cite{kim2022flame} introduces a groundbreaking transformer-based architecture that effectively manages motion data, which is crucial for handling variable-length motions and attending to free-form text. ~\cite{chen2022executing} presents a Motion Latent-based Diffusion model (MLD) that generates plausible human motion sequences based on various conditional inputs, such as action classes or textual descriptors. ~\cite{tevet2022human} proposes the Motion Diffusion Model (MDM), a meticulously adapted classifier-free diffusion-based generative model for the human motion domain. Building on this work,~\cite{yuan2022physdiff} integrates physical constraints into the diffusion process to generate physically plausible human motions.
{Another research line is to employ autoregressive models equipped with motion quantization (VQ-VAE). This category of methods first tokenizes the continuous motion into discrete motion tokens and then models the sequence using various autoregressive models~\citep{guo2022tm2t,zhang2023t2m,lucas2022posegpt}, even the pre-trained LLMs~\citep{kalakonda2023action,zhang2023motiongpt,jiang2023motiongpt}.}

The aforementioned motion generation techniques primarily focus on single-person motion generation, lacking the capability to generate and model dual or multi-person human interaction motions. Recent work by ~\cite{shafir2023human} suggests that the gap in data availability for motion generation can be addressed using a pre-trained diffusion-based model~\citep{tevet2022human} as a generative prior and demonstrates the prior's effectiveness for fine-tuning in a few-shot manner. However, this approach is limited to generating interactions observed during training, resulting in constrained generalization and suboptimal generation quality capabilities. 
{A concurrent work RIG by \cite{tanaka2023role} attempts to recover 3D interaction motions from a noisy depth dataset~\citep{liu2019ntu} and translates the motion labels into sentences. 
However, this approach yields motion data that is often unrealistic and text annotations with poor matching quality. Although It also coincidentally models the symmetry of two-person interaction, the limited data quality and quantity impede further exploration of the problem. 
It continues to face several challenges, including drifts, foot-sliding, unnatural interactions, and limited generalization to novel texts.}

\begin{table*}[tbp]

	\begin{center}
		\centering
		\caption{\textbf{Dataset comparisons.} We compare our InterHuman dataset with existing human motion datasets. \textbf{Motions} refers to the total number of motion clips. \textbf{Vocab.} shows the number of distinct words used in the annotations, while \textbf{Descriptions} summarizes the total number of textual descriptions.}\label{Tab.1}
		\resizebox{1\textwidth}{!}{
		\begin{tabular}{lccccccccc}
			\toprule
			Dataset     &Natural Language    &Interactive    &Motions         &Vocab.         &Descriptions        & Duration         \\
			\midrule
			KIT \citep{Plappert2016kit}     &$\checkmark$   &-    & 3911    & 1623    & 6278       & 11.23h    \\

			HumanML3D \citep{guo2022generating}   &$\checkmark$    &-    & 14,616   & 5371   &44,970       & 28.59h    \\
			
                BABEL \citep{punnakkal2021babel}       &-    & -   & 13,220   & -   & -           & 43.5h \\	
                {ExPI \citep{guo2022multi}}       &-    & $\checkmark$   & 115   & -   & -           & 0.28h \\	
                NTU RGB+D 120 \citep{liu2019ntu}       &-    &$\checkmark$    & 20,579   & -   & -           & 18.6h \\	
            UMPM \citep{van2011umpm} &-     &$\checkmark$    & 36    & -    & -  & 2.22h \\
			
			You2Me \citep{ng2020you2me}         &-     &$\checkmark$    & 42    & -    & -     & 1.4h      \\
			\midrule
			\textbf{InterHuman(Ours)}   &\textbf{$\checkmark$}    &\textbf{$\checkmark$}    & \textbf{7779}    &\textbf{5656}  &\textbf{23,337}     &\textbf{6.56h}       \\
			
			\bottomrule
		\end{tabular}
		}
	\end{center}
 
\end{table*}

\subsection{Human Motion Capture}\label{human motion capture}

Motion capture techniques have been well-developed in the last decade.
Marker-based techniques, such as~\cite{VICON} and those presented by~\cite{vlasic2007practical}, have been successful in capturing high-quality human motions for professional applications. However, these methods are not suitable for daily use due to their costly and laborious setup. To overcome this limitation, markerless motion capture methods have been developed~\citep{bregler1998tracking,de2008performance,theobalt2010performance}. 
Advances in parametric human models~\citep{anguelov2005scape,loper2015smpl,pavlakos2019expressive,osman2020star} have led to data-driven approaches for estimating 3D human pose and shape using optimization~\citep{huang2017towards,lassner2017unite,bogo2016keep,kolotouros2019learning} or direct regression~\citep{kanazawa2019learning,kocabas2020vibe,zanfir2020neural} of human model parameters. Template-based methods, utilizing specific template meshes as priors, have been proposed for both multi-view~\citep{Gall2010,StollHGST2011,liu2013markerless,robertini2016model,Pavlakos17,Simon17,FlyCap} and monocular~\citep{MonoPerfCap,LiveCap2019tog,EventCap_CVPR2020,DeepCap_CVPR2020} setups.
Another research line is inertial measurement units (IMUs). Commercially available systems, such as Xsens MVN~\citep{XSENS}, have employed large numbers of sensors, but the intrusive nature of these systems has prompted research into sparser sensor setups.~\cite{von2017SIP} presented a pioneering exploration called SIP, which employs only six IMUs. However, the traditional optimization framework used in SIP hampers real-time application. Data-driven approaches~\citep{huang2018DIP,TransPose2021,PIPCVPR2022} utilizing sparse sensors have shown significant improvements in accuracy and efficiency, but substantial drift remains an issue for challenging motions. Previous sensor-aided solutions have combined IMUs with videos ~\citep{gilbert2019fusing,henschel2020accurate,malleson2019real,malleson2017real,liang2022hybridcap}, RGB-D cameras~\citep{helten2013real,Zheng2018HybridFusion}, optical markers~\citep{Andrews2016}, or even LiDAR~\citep{zhao2022lidar} to address the scene-occlusion problem and effectively correct drift. 
These methods achieve highly accurate capture of human motions given various modalities of signals, however, the high-level control of synthesizing captured motion and even non-seen motions with more modalities of input remains challenges.

\subsection{Human Motion Dataset}\label{human motion dataset}
In recent times, the accessibility of extensive motion datasets has played a crucial role in propelling motion generation research forward.
Action label datasets such as BABEL~\citep{punnakkal2021babel}  and NTU RGB+D 120~\citep{liu2019ntu}, although having labeled annotations, differ significantly from natural language when composed of verb or verb-object structures, and furthermore, the action labels only enable direct classification of actions, making it impossible to support text-based human motion generation. Datasets such as KIT~\citep{Plappert2016kit}, and HumanML3D~\citep{guo2022generating}, which feature text annotations, have been especially valuable for the progression of text-driven motion generation research~\citep{chen2022executing,tevet2022human,petrovich2022temos}. However, these datasets only consist of single-person motions and annotations, making it difficult to apply and generalize to the generation of interactive motions involving two or more individuals.

Various multi-person motion datasets have been developed, including
3DPW~\citep{von2018recovering}, You2Me~\citep{ng2020you2me}, and UMPM~\citep{van2011umpm}. However, while these datasets contain two-person and multi-person motion data, they are limited in size and annotations. In particular, there is a lack of textual or other modal annotations in these datasets.
Efforts to annotate existing datasets with text, as demonstrated in the annotation of 3DPW~\citep{von2018recovering} in ComMDM~\citep{shafir2023human}, establish a foundation for future advancements in text-guided multi-person motion generation. However, since it only contains 27 two-person motion sequences, the issue of limited availability of two-person interaction datasets still persists. Our proposed InterHuman dataset is currently the largest interaction-language dataset, addressing the lack of suitable datasets in text-based human-to-human interactive motion generation research.

\section{InterHuman Dataset}\label{interhuman dataset}

InterHuman\footnote{The captured skeletal motions and text annotations are available at \url{https://tr3e.github.io/intergen-page/}.} is a comprehensive, large-scale 3D human interactive motion dataset encompassing a diverse range of 3D motions of two interactive people, each accompanied by natural language annotations. To the best of our knowledge, it is the most extensive 3D human-to-human interaction dataset available. Unlike some previous datasets that focus only on single-person motion or particular actions, such as dancing~\citep{li2021ai}. Our dataset comprises various interaction motions, broadly classified into two categories: daily motion, which encompasses everyday routines involving two people (e.g., passing objects, greeting, communicating, etc.), and professional motions, which include typical human-to-human interactions (e.g., Taekwondo, Latin dance, boxing, etc.). 

\subsection{Data collection}\label{data collection}
We utilized a motion capture system with 76 calibrated Z-CAM~\citep{Z-CAM} RGB cameras to capture human interaction motions, the Fig.~\ref{Fig.2} shows our system and captured motion data. 

{Our dataset was amassed through two distinct sessions: daily motion and professional motion. The former encompasses a spectrum of high-frequency interactions encountered in everyday life, while the latter is tailored to capture professional interaction performances, consisting of 10 specific categories of expert skills shown in Fig.~\ref{Fig.4}. This bifurcated approach to data collection was meticulously designed to cover the motion distribution of real-world interactions as comprehensively as possible.}
{\textbf{In the realm of daily motion}, we first involved 15 adept scriptwriters without communication with each other, each charged with the task of crafting textual scripts that vividly depict a wide array of two-person interaction scenarios, thereby ensuring maximal diversity.
Subsequently, we engaged 18 pairs of drama actors, each pair possessing a rapport, to bring these text-scripted interactions to real motions through their performances.
\textbf{In the realm of professional motion}, we engaged 12 pairs of performers, each hailing from diverse disciplines and possessing unique skill sets, such as dance and martial arts. This included two pairs specializing in Latin dance and two pairs in taekwondo, with one pair representing each of the remaining categories. To ensure motion diversity, the performers, already well-acquainted with their respective partners, were encouraged to exhibit a broad spectrum of interactions based on their own collaborative experiences during their performances.}

After that, we adopted a data processing pipeline analogous to the one described by \citet{li2021ai} to extract SMPL parameters (Loper et al., 2015) from the captured multi-view videos. Subsequently, we initiated a textual annotation endeavor utilizing Amazon Mechanical Turk (AMT). {During this phase, we instructed annotators to segment the videos into the most reasonable discrete clips to maximally preserve the semantic meaning of each interaction, each with a maximum duration of 10 seconds. We then obtained three distinct textual descriptions for every clip from separate annotators.} Illustrative examples of these annotations are depicted in Fig.~\ref{Fig.3}, where the detailed interactions between the two people are described from different perspectives.

\subsection{Dataset comparison}\label{data comparison}
\begin{figure}[tbp] 

	\centering  
	\includegraphics[width=1.0\linewidth]{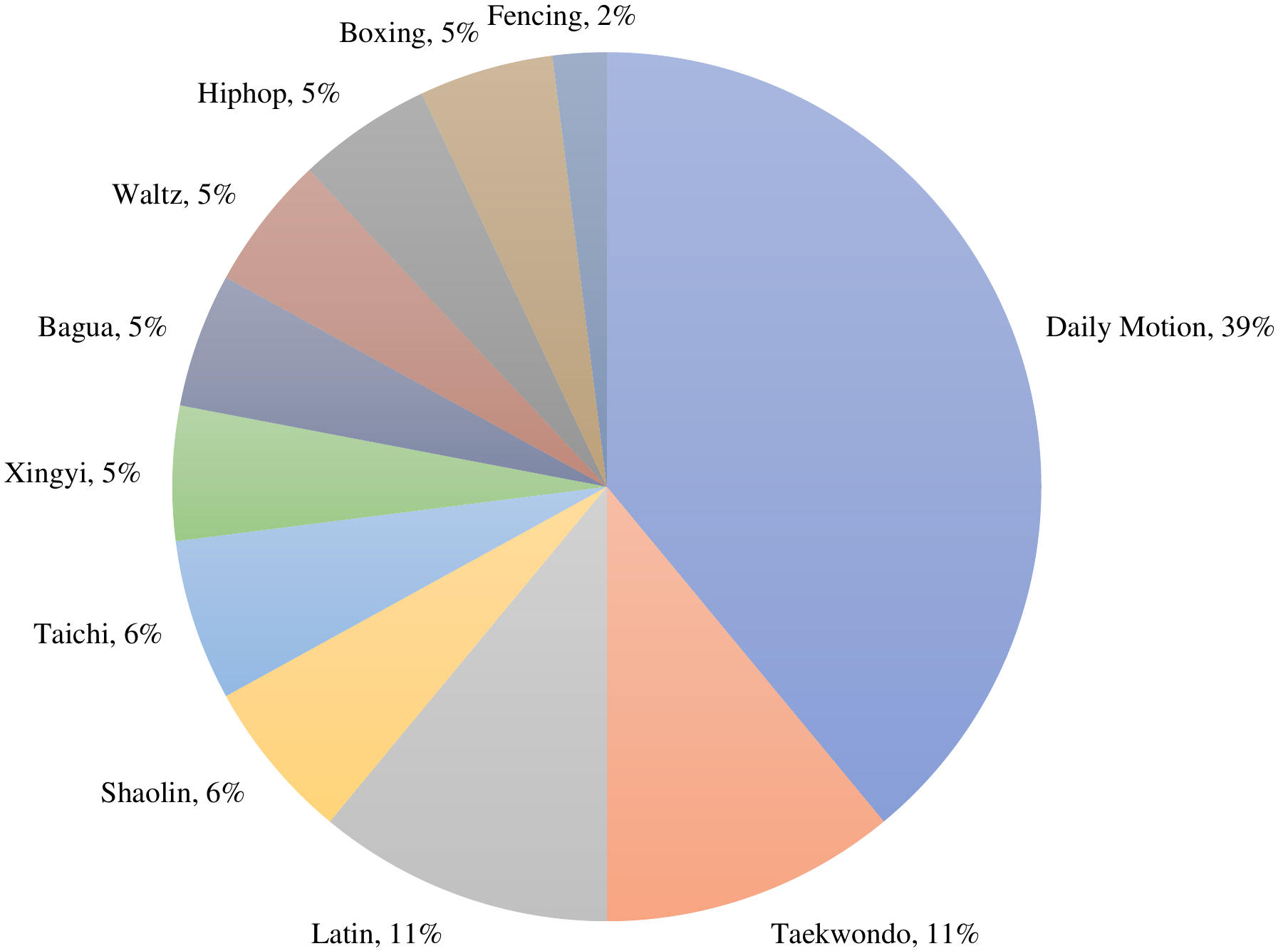} 
	\caption{InterHuman dataset covers a wide range of two-
person interactions, from the daily ones like hugging, handshake, and argument to
the professional motions ranging from dance to martial arts.} 
	
 \label{Fig.4} 
\end{figure}

Our InterHuman dataset was captured using a system with 76 cameras, and all motions have been meticulously annotated, consisting of 7779 motions derived from various categories of human actions, labeled with 23,337 unique descriptions composed of 5656 distinct words, with a total duration of 6.56 hours, which makes it the largest and most diverse known scripted dataset of human-to-human interactions. For specific durations of each motion category, please refer to Fig.~\ref{Fig.4}. Tab.~\ref{Tab.1} provides a comprehensive comparison of our dataset with several existing human datasets from various perspectives, highlighting that our dataset is currently the most suitable for tasks involving human interactions.

\begin{figure*}[ht] 
	\centering  
	\includegraphics[width=1.0\linewidth, height=0.46\linewidth]{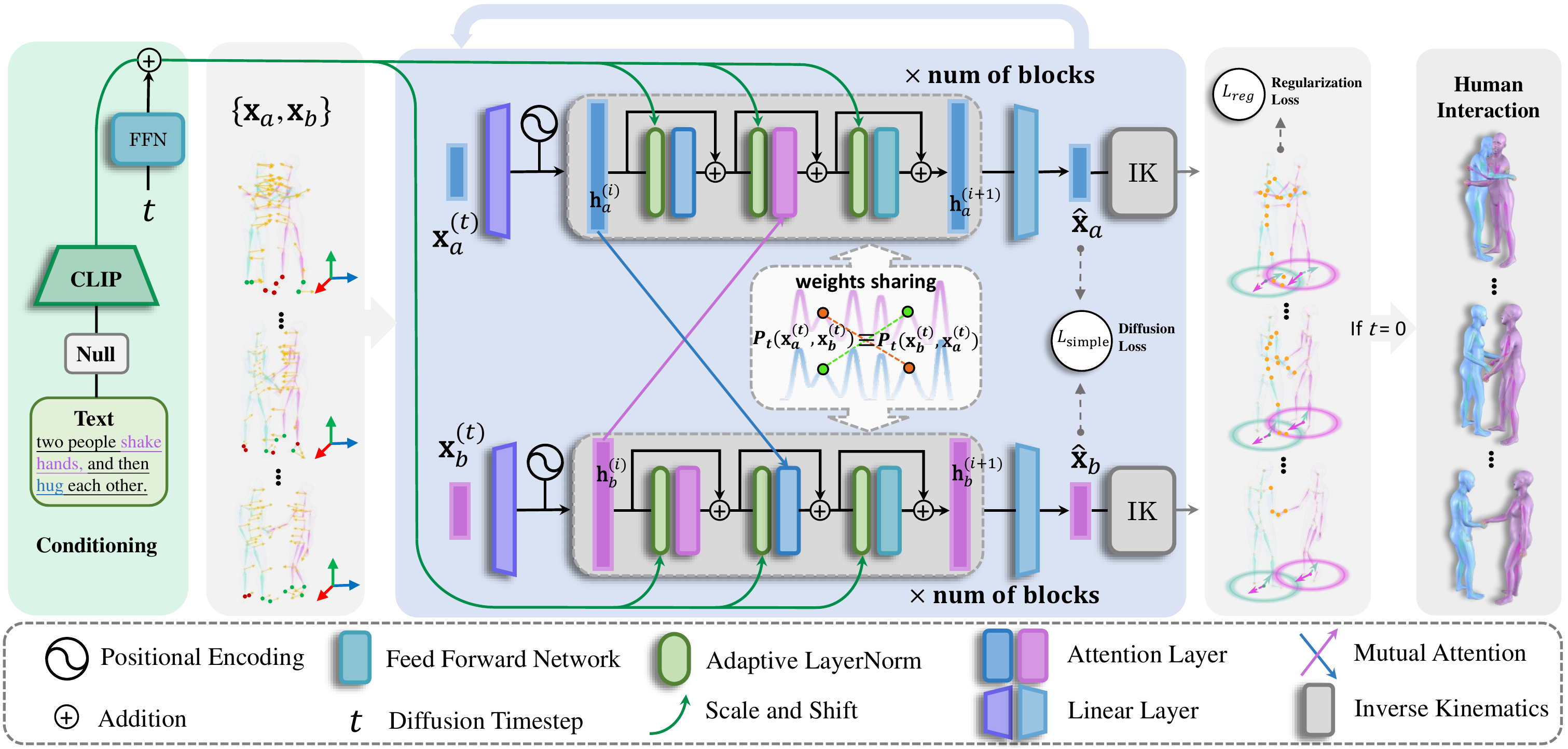} 
	\caption{\textbf{The overview of our InterGen.} We contribute three primary technical designs. First, we propose an efficient two-person interaction motion representation. Second, we introduce two cooperative transformer-style weights-sharing networks with mutual attention to interactively perform denoising. Lastly, we introduce an effective loss function that significantly improves the quality of two-person interaction generation.} 
	
 \label{Fig.5} 
\end{figure*} 

\section{InterGen Approach}\label{intergen approach}

Our goal is to generate diverse and high-quality human interaction motions conditioning on text prompts, within a diffusion-based framework. 
To this end, as illustrated in Fig.~\ref{Fig.5}, our approach consists of three key technical designs. 
The first is an effective motion representation (Sec.~\ref{Sec.4.1}) that preserves the spatial relations of interacting people in the common world frame.
Then, we adopt a novel denoising architecture that involves two cooperative networks (Sec.~\ref{Sec.4.2}) with sharing-weights and mutual attention to connect the two branches at hierarchical feature levels, so as to balance the motion capacity of two interacting performers. 
To train such interaction diffusion model, we further propose additional regularization terms (Sec.~\ref{Sec.4.3}), consisting of a joint distance map (DM) loss and a relative orientation (RO) loss to enforce the networks to depend on each others, especially under continuous and close interactions. In addition, we propose a novel loss damping schedule during training to improvive the generation results.

\subsection{Human Interaction Representation} \label{Sec.4.1}

We model a two-person interaction $\textbf{x}$ as a collection of two single-person motion sequences $\textbf{x}_h$, i.e., $\textbf{x}=\{\textbf{x}_a, \textbf{x}_b\}$, where $\textbf{x}_h=\{x^i\}^L_{i=1}$ is a fixed-framerate sequence of motion states $x^i$. Thus these two sequences are naturally synchronized. The core problem is to encode the spatial relationships between them.

\myparagraph{Canonical representation.}\label{canonical representation}
HumanML3D~\citep{guo2022generating} proposed a human motion representation for single-person scenarios that incorporate ground contact information and motion features. This representation is over-parameterized, expressive, and neural network friendly, and has been adopted by several recent works. However, this representation cannot be directly applied to multi-person scenarios because it canonicalizes joint positions and velocities to the root frame, which loses global spatial information.
ComMDM~\citep{shafir2023human} tries to mitigate this issue by predicting the initial relative rotation and translation between two people. In contrast, we extend this representation by introducing global relative rotation $\textbf{r}^h\in \mathbb{R}^2$ along Y-axis and translation $\textbf{t}^h\in \mathbb{R}^2$ on XZ-plane over other humans.
\begin{align}
x^i=[\dot{r}^a , \dot{r}^x, \dot{r}^z, r^y, \textbf{j}^p_{l} , \textbf{j}^v_{l} , \textbf{j}^r, \textbf{c}^f, \textbf{r}^h, \textbf{t}^h ],
\end{align}
where the $i$-th motion state $x^i$ is defined as a collection of root angular velocity $\dot{r}^a\in \mathbb{R}$ along the Y-axis, root linear velocities $\dot{r}^x, \dot{r}^z\in \mathbb{R}$ on the XZ-plane, root height $r^y \in \mathbb{R}$, local joint positions $\textbf{j}^p_{l} \in \mathbb{R}^{3N_j}$, local velocities $\textbf{j}^v_{l} \in \mathbb{R}^{3N_j}$, rotations $\textbf{j}^r \in \mathbb{R}^{6N_j}$ in root space, and binary foot-ground contact features $\textbf{c}^f \in \mathbb{R}^{4}$ by thresholding the heel and toe joint velocities, where $N_j$ denotes the joint number.

\myparagraph{Non-canonical representation.}\label{non-canonical representation}
{Nevertheless, the global absolute trajectories derived from canonical representation suffer from drifts, owing to the compulsory cumulative integration of noisy local velocities. This deficiency precipitates an accumulation of errors, culminating in unbounded exponential drifts over time, as delineated by \cite{von2017SIP}.
This absolute trajectory error can be ignored for tasks such as single-person short-sequence motion synthesis, which focuses on the plausibility of local motion and does not care much about the absolute trajectory. Whereas it is fatal for tasks such as multi-person motion synthesis, which require person-to-person precise spatial relations.}

{To this end, we propose a non-canonical representation for multi-person interaction motion. 
Our key idea is to incorporate the absolute trajectories of both individuals within the same world frame directly into the motion representation, thereby bypassing the cumulative integration process.}
Instead of transforming joint positions and velocities to the root frame, we keep them in the world frame, which allows us to directly access the global translation and rotation from the root position and inverse kinematics (IK), respectively, to avoid drift effectively. The representation is formulated as:
\begin{align}
x^i=[\textbf{j}^{p}_{g} , \textbf{j}^{v}_{g} , \textbf{j}^r, \textbf{c}^f ],
\end{align}
where the $i$-th motion state $x^i$ is defined as a collection of global joint positions $\textbf{j}^{p}_{g}\in \mathbb{R}^{3N_j}$, velocities $\textbf{j}^{v}_{g}\in \mathbb{R}^{3N_j}$ in the world frame, 6D representation of local rotations $\textbf{j}^{r}\in \mathbb{R}^{6N_j}$ in the root frame, and binary foot-ground contact features $\textbf{c}^f \in \mathbb{R}^{4}$.

\subsection{Human Interaction Diffusion}  \label{Sec.4.2}

\myparagraph{Diffusion models.}\label{diffusion models}
We sometimes omit the explicit dependence on condition c for simplicity of notation. Note that we can always train diffusion models with some condition $c$; even for the unconditional case, we can condition the model on a universal null token $\emptyset$.

Let $p_0(\textbf{x})$ denote the human interactive motion data distribution, diffusion models~\cite{ho2020denoising} injects time-dependent $i.i.d$ Gaussian noise to the samples from $p_0$, giving a diffusion process $\{p_t(\textbf{x})\}_{t=0}^T$ with a continuous variable $t \in [0, T]$.
Then a generative model can be obtained by reversing the process, starting from samples $\textbf{x}^{(T)} \thicksim p_T$ that is a standard Gaussian distribution, and then solving the following reverse-time SDE from $t=T$ to $t=0$:
\begin{align}
d\textbf{x} = [\textbf{f}(\textbf{x}, t)-\sigma_t^2\nabla_\textbf{x} log p_t(\textbf{x})]dt + \sigma_t d\textbf{w},
\end{align}
where $\textbf{f}(\cdot, t)$ is a deterministic drift function, $\sigma_t$ is the diffusion coefficient that is increasing over time to control the noise level, $dt$ is infinitesimal negative timestep, $d\textbf{w}$ is infinitesimal noise, and $\nabla_\textbf{x}log p_t(\textbf{x})$ is the score function that is the only intractable term. 
Note that it can be obtained from the expectation of $\textbf{x}$ given $\textbf{x}^{(t)}$:
\begin{align}
\nabla_{\textbf{x}^{(t)}} log p_t(\textbf{x}^{(t)}) = (\mathbb{E} [\textbf{x}|\textbf{x}^{(t)}] - \textbf{x}^{(t)})/\sigma_t^2.
\end{align}
When we drop the noise term at Eq.~(3), an ordinary differential equation (ODE) is obtained, which is a corresponding deterministic process sharing the same marginal distribution $\{p_t(\textbf{x})\}_{t=0}^T$ and is referred to as probability flow ODE, which can accelerate sampling process by performing a linear interpolation between $\textbf{x}^{(t)}$ and the predicted $\mathbb{E} [\textbf{x}|\textbf{x}^{(t)}]$~\citep{song2020denoising}.

\myparagraph{Interaction Diffusion.}\label{interaction diffusion}
Our approach is based on a fundamental assumption, commutative property, which means that two-person interactions $\{\textbf{x}_a, \textbf{x}_b\}$ and $\{\textbf{x}_b, \textbf{x}_a\}$ are equivalent, i.e., the order of every single motion does not change the semantics of the interaction itself. In other words, the distribution of interaction data satisfies the following property: 
\begin{align}
p(\textbf{x}_a, \textbf{x}_b) \equiv p(\textbf{x}_b, \textbf{x}_a). 
\end{align}
Under this assumption, the two people share the same single-person motion marginal distribution. Since noise $\epsilon \sim \mathcal{N}(\mathbf{0},\mathbf{I})$ is independent of data distribution, we have:
\begin{align}
p_t(\textbf{x}_a^{(t)}, \textbf{x}_b^{(t)}) \equiv p_t(\textbf{x}_b^{(t)}, \textbf{x}_a^{(t)}). 
\end{align}
Thus based on the above conclusion the score function  can be reformulated as:
\begin{align}
&\nabla_{\textbf{x}^{(t)}} log p_t(\textbf{x}^{(t)}) \nonumber\\ 
& = [\nabla_{\textbf{x}_a^{(t)}} log p_t({\textbf{x}_a}^{(t)}, {\textbf{x}_b}^{(t)}), \nabla_{\textbf{x}_b^{(t)}} log p_t({\textbf{x}_a}^{(t)},{\textbf{x}_b}^{(t)})] \nonumber\\ 
& = [\nabla_{\textbf{x}_a^{(t)}} log p_t({\textbf{x}_a}^{(t)}, {\textbf{x}_b}^{(t)}), \nabla_{\textbf{x}_b^{(t)}} log p_t({\textbf{x}_b}^{(t)},{\textbf{x}_a}^{(t)})],
\end{align}
where the two parts for $\textbf{x}_a$ and $\textbf{x}_b$ are the same function $\nabla_{a} log p_t(a, b)$, which can be approximated by employing the same network with the following denoising autoencoder objective:
\begin{align}
\mathcal{L}_{simple} = 
\mathbb{E}_{\textbf{x}, t, \epsilon} [&\lambda_t  ||\textbf{x}_a -D_{\theta}(\textbf{x}_a+\sigma_t \epsilon_a, \textbf{x}_b+\sigma_t \epsilon_b, t, c)||_2^2 \nonumber\ \\
+ &\lambda_t  ||\textbf{x}_b -D_{\theta}(\textbf{x}_b+\sigma_t \epsilon_b, \textbf{x}_a+\sigma_t \epsilon_a, t, c)||_2^2],
\end{align}
 where $D_{\theta}$ is the denoisers sharing the common network weights, whose input consists of its own noisy motion to denoise, the cooperator's noisy motion, the time $t$, and the condition $c$, noise $\epsilon \sim \mathcal{N}(\mathbf{0},\mathbf{I})$, and $\lambda_t$ is the loss weighting factor.

\myparagraph{Cooperative denoisers.}\label{cooperative denoisers}
Based on the aforementioned conclusions, we adopt interactively cooperative transformer-style networks sharing the common weights to model $D_\theta$, as demonstrated in Fig.~\ref{Fig.5}. The networks are fed their own noisy motions, $\textbf{x}_a^{(t)}$ and $\textbf{x}_b^{(t)}$, as inputs for denoising and subsequently output the corresponding denoised versions of the motions, $\textbf{x}_a$ and $\textbf{x}_b$. This prediction process is conditioned on the diffusion timestep $t$, control condition $c$, and the hidden states $\textbf{h}^{(i)}$ of the counterpart network.

Specifically, the noisy motion is first embedded into a common latent space and positionally encoded into an internal representation often referred to as the hidden states $\textbf{h}^{(0)}$. Then, it is processed by $N$ attention-based blocks to obtain denoised hidden states $\textbf{h}^{(N)}$. Finally, a common inverse embedding layer is applied to output the denoised motion.

Each block consists of two multi-head attention layers ($Attn$) followed by one feed-forward network ($FF$). 
The first attention layer is a self-attention layer, which embeds the current hidden states $\textbf{h}^{(i)}$ into a context vector $\textbf{c}^{(i)}$. The computation of $\textbf{c}_a^{(i)}$ part is formulated as the following:
\begin{align}
&\textbf{c}_a^{(i)} = Attn(\textbf Q,\textbf K,\textbf V)
=softmax(\frac{\textbf Q \textbf K^T}{\sqrt{C}}) \textbf V, \nonumber \\
&\textbf Q = \textbf{h}_a^{(i)} \textbf W_s^{Q}, \textbf K = \textbf{h}_a^{(i)} \textbf W_s^{K}, \textbf V = \textbf{h}_a^{(i)} \textbf W_s^{V},
\end{align}
where $C$ is the number of channels in the attention layer and $\textbf W_s$ are trainable weights, and $\textbf{c}_b^{(i)}$ is calculated with shared weights $\textbf W_s$ in the same way.

The second attention layer is a mutual attention layer, where the key $\textbf K$ value $\textbf V$ pair is provided by the hidden states $\textbf{h}^{(i)}$ of the counterpart block. Here the computation process of the next hidden states $\textbf{h}^{(i+1)}$ is formulated as:
\begin{align}
&\textbf{h}_a^{(i+1)} = FF(Attn(\textbf Q_a,\textbf K_b,\textbf V_b)), \nonumber \\
&\textbf{h}_b^{(i+1)} = FF(Attn(\textbf Q_b,\textbf K_a,\textbf V_a)), \nonumber \\
&\textbf Q_a = \textbf{c}_a^{(i)} \textbf W_m^{Q}, \textbf K_a = \textbf{h}_a^{(i)} \textbf W_m^{K}, \textbf V_a = \textbf{h}_a^{(i)} \textbf W_m^{V}, \nonumber \\
&\textbf Q_b = \textbf{c}_b^{(i)} \textbf W_m^{Q}, \textbf K_b = \textbf{h}_b^{(i)} \textbf W_m^{K}, \textbf V_b = \textbf{h}_b^{(i)} \textbf W_m^{V},
\end{align}
where $\textbf W_m$ are trainable weights shared by the two branches.

In addition, the adaptive layer normalization is employed before all attention layers and the feed-forward network to condition on the control condition $c$ and timestep $t$.

\subsection{Additional Regularization Losses} \label{Sec.4.3}

\myparagraph{Geometric losses.}\label{geometric losses}
We adopt the common geometric losses in the field of human motion, such as foot contact loss $\mathcal{L}_{foot}$ and joint velocity loss $\mathcal{L}_{vel}$, to regularize the generative models and enforce physical plausibility and coherence for each single-person motion. For more details, we refer the reader to MDM~\citep{tevet2022human}. In addition, for our non-canonical representation, we introduce bone length loss $\mathcal{L}_{BL}$ to constrain the global joint positions of each person to satisfy skeleton consistency, which implicitly encodes the human body’s kinematic structure. We formulate the bone length loss as follows:
\begin{align}
\mathcal{L}_{BL} = ||B(\hat{\textbf{x}}_a) - B(\textbf{x}_a)||_2^2 + ||B(\hat{\textbf{x}}_b) - B( \textbf{x}_b)||_2^2 ,
\end{align}
where $B$ represents the bone lengths in a pre-defined human body kinematic tree, derived from the global joint positions in $\textbf{x}$.

\begin{figure}[tbp] 
	\centering  
	\includegraphics[width=1.0\linewidth]{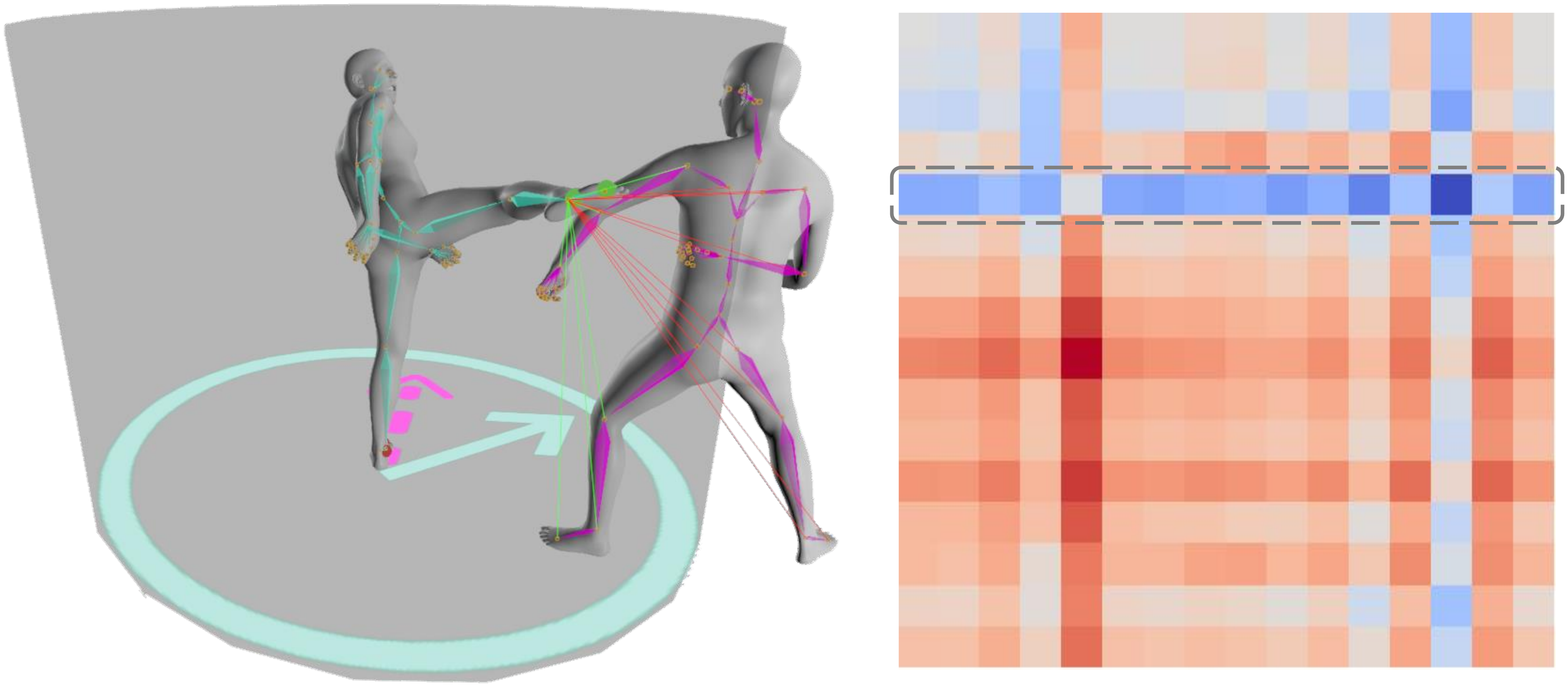} 
	\caption{(Left) visualize our proposed interactive losses, where the relative orientation loss is the angular separation between the frontal orientations of the two people. And the partial joint distance map of the heel joint is truncated with the region of the cylinder, which is shown in (right), where the highlighted row encodes the spatial relations between the heel and the other person.} 
 \label{Fig.6} 
\end{figure}

\myparagraph{Interactive losses.}\label{interactive losses}
To handle the complexity of spatial relations in multi-person interactions, we further introduce interactive losses, comprising masked joint distance map (DM) loss and relative orientation (RO) loss, as illustrated in Fig.~\ref{Fig.6}. 
The DM loss measures the $N_j\times N_j$ joint distance map of two people and matches it with the ground truth, where $N_j$ is the number of joints per person. Thus we design the DM loss as follows:
\begin{align}
& \mathcal{L}_{DM} \nonumber \\
&  =||(M(\hat{\textbf{x}}_a, \hat{\textbf{x}}_b) - M(\textbf{x}_a, \textbf{x}_b)) \odot I(M_{xz}(\textbf{x}_a, \textbf{x}_b) < \bar{M})||_2^2,
\end{align}
where $M$ denotes the joint distance map of two people, obtained from the global joint positions in their motions, $I(\cdot)$ is the indicator function that masks the loss by applying a 2D distance threshold on the XZ-plane, which activates this loss only when the horizontal distance between the two people is small enough, $M_{xz}$ represents the distance map projected onto the XZ-plane, $\bar M$ is the distance threshold, and $\odot$ indicates Hadamard product.

\begin{figure*}[tbp] 
	\centering  
	\includegraphics[width=1.0\linewidth]{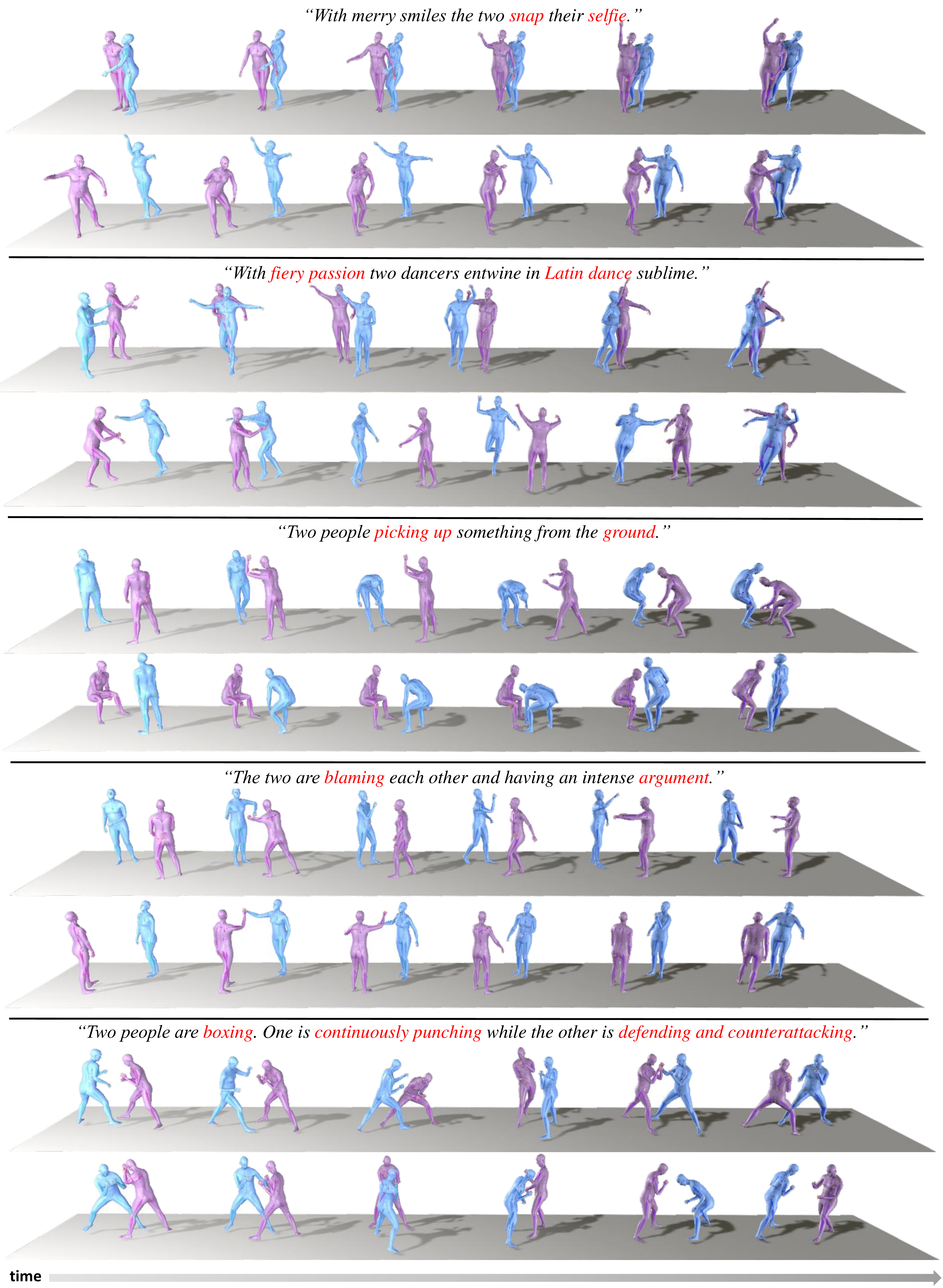} 
	\caption{\textbf{Qualitative results} generated by our InterGen model. We showcase two different samples per text prompt, which demonstrate the high quality and diversity of our interaction generation.} 
	
 \label{Fig.7} 
\end{figure*}

\begin{table*}[tbp]

	\begin{center}
		\centering
		\caption{\textbf{Quantitative comparisons on the InterHuman test set.} We run all the evaluations 20 times except MModality runs 5 times. $\pm$ indicates the 95\% confidence interval. \textbf{Bold} indicates best result. 
{ComMDM* indicates the ComMDM model fine-tuned in the original few-shot setting with 10 training samples and ComMDM (without *) indicates fine-tuned on our entire InterHuman training set. All the models employ the same non-canonical representation.}} 
        \label{Tab.2}
		
 		\resizebox{1\textwidth}{!}{
        		\begin{tabular}{lccccccc}
        		    \toprule
        		    \multirow{2}{*}{Methods}  & \multicolumn{3}{c}{R Precision$\uparrow$} & \multirow{2}{*}{FID $\downarrow$} & \multirow{2}{*}{MM Dist$\downarrow$}  & \multirow{2}{*}{Diversity$\rightarrow $} & \multirow{2}{*}{MModality $\uparrow$}\\
        		    \cmidrule(lr){2-4}
        			& Top 1 & Top 2  & Top 3 \\
        			\midrule
                        Real    &  $0.452^{\pm .008}$ &  $0.610^{\pm .009}$  & $0.701^{\pm .008}$   &  $0.273^{\pm .007}$ &  $3.755^{\pm .008}$  & $7.948^{\pm .064}$   & -  \\
                        \midrule
        			TEMOS {\citep{petrovich2022temos}}  &  $0.224^{\pm .010}$ &  $0.316^{\pm .013}$  & $0.450^{\pm .018}$   &  $17.375^{\pm .043}$ &  $6.342^{\pm .015}$  & $6.939^{\pm .071}$   & $0.535^{\pm .014}$  \\
        			T2M {\citep{guo2022generating}}  &  $0.238^{\pm .012}$ &  $0.325^{\pm .010}$  & $0.464^{\pm .014}$   &  $13.769^{\pm .072}$ &  $5.731^{\pm .013}$  & $7.046^{\pm .022}$   & $1.387^{\pm .076}$ \\
        			MDM {\citep{tevet2022human}}   &  $0.153^{\pm .012}$ &  $0.260^{\pm .009}$  & $0.339^{\pm .012}$   &  $9.167^{\pm .056}$ &  $7.125^{\pm .018}$  & $\textbf{7.602}^{\pm .045}$   & $\textbf{2.355}^{\pm .080}$ \\
                    ComMDM$^*$ {\citep{shafir2023human}}  &  $0.067^{\pm .013}$ &  $0.125^{\pm .018}$  & $0.184^{\pm .015}$   &  $38.643^{\pm .098}$ &  $14.211^{\pm .013}$  & $3.520^{\pm .058}$   & $0.217^{\pm .018}$  \\
                    ComMDM {\citep{shafir2023human}}  &  $0.223^{\pm .009}$ &  $0.334^{\pm .008}$  & $0.466^{\pm .010}$   &  $7.069^{\pm .054}$ &  $6.212^{\pm .021}$  & $7.244^{\pm .038}$   &  $1.822^{\pm .052}$ \\
                    {RIG \citep{tanaka2023role}}  &  {$0.285^{\pm .010}$} &  {$0.409^{\pm .014}$}  & {$0.521^{\pm .013}$}   &  {$6.775^{\pm .069}$} &  {$5.876^{\pm .018}$}  & {$7.311^{\pm .043}$}   & {$2.096^{\pm .065}$}   \\
        			\midrule
                    InterGen (ours)    &  $\textbf{0.371}^{\pm .010}$ &  $\textbf{0.515}^{\pm .012}$  & $\textbf{0.624}^{\pm .010}$   &  $\textbf{5.918}^{\pm .079}$ &  $\textbf{5.108}^{\pm .014}$  & $7.387^{\pm .029}$ & $2.141^{\pm .063}$ \\
        			\bottomrule
        		\end{tabular}
 		}
	\end{center}

\end{table*}

The RO loss estimates the relative orientation of two people and aligns it with the ground truth.  The RO loss is formulated as:
\begin{align}
&\mathcal{L}_{RO} = ||O(IK(\hat{\textbf{x}}_a), IK(\hat{\textbf{x}}_b)) - O(IK(\textbf{x}_a), IK(\textbf{x}_b))||_2^2,
\end{align}
where $IK(\cdot)$ represents the inverse kinematics process, which outputs the joint rotations, and $O$ indicates the 2D relative orientation between the two people around the Y-axis obtained from rotations.
These losses constrain the positional spatial relations and the relative frontal orientation of the two people to be consistent with the nature of human interactions.

Here, we summarize our additional regularization loss as follows:
\begin{align}\label{eqn:regloss}
\mathcal{L}_{reg} = &\lambda_{vel} \mathcal{L}_{vel} + \lambda_{foot} \mathcal{L}_{foot} + \lambda_{BL} \mathcal{L}_{BL}  \nonumber \\
&+ \lambda_{DM} \mathcal{L}_{DM} + \lambda_{RO} \mathcal{L}_{RO},
\end{align}
{where the hyper-parameters $\lambda_{vel}, \lambda_{foot}, \lambda_{BL}, \lambda_{DM}, \lambda_{RO}$ are meticulously calibrated to regulate the magnitude orders of their corresponding terms. This calibration is effective in addressing the existing disparities in magnitude orders among various loss terms. This harmonizes distinct loss terms into the same magnitude order, thereby ensuring a balanced contribution from each term.}

\myparagraph{Regularization loss schedule.}\label{regularization loss schedule}
Physdiff~\citep{yuan2022physdiff} informed that the network-predicted denoised motion is increasingly implausible as the diffusion timestep $t$ is larger (noise level is higher). 
Since denoisers estimate the expectation $\mathbb{E} [\textbf{x}|\textbf{x}^{(t)}]$, i.e., the mean motion $\textbf{x}$ given $\textbf{x}^{(t)}$, and empirically they tend to output average poses with some root translations and rotations when the noise level is high. 
This results in not only severe physical implausible motions but also unrealistic interactions between two people in our two-person scenario.
If we apply the above regularization losses when $t$ is large, the network output will become the minimum mean squared error (MMSE) estimation of biased losses, deviating from $\mathbb{E} [\textbf{x}|\textbf{x}^{(t)}]$.

Inspired by that, we devise a novel diffusion training scheme. We truncate diffusion timesteps with a threshold $\bar t$ and only apply regularization loss to the network when the sampled timestep $t$ is below the threshold. Thus the total loss is formulated as:
\begin{align}\label{eqn:loss}
\mathcal{L} = \mathcal{L}_{simple} + \lambda_{reg} \mathbb{E}_t [I(t \leq \bar t) \cdot \mathcal{L}_{reg}^{(t)}],
\end{align}
where $I(t \leq \bar t)$ is an indicator function, which drops the regularization term when $t > \bar t$.

\begin{figure*}[htbp] 
	\centering  
	\includegraphics[width=0.92\linewidth]{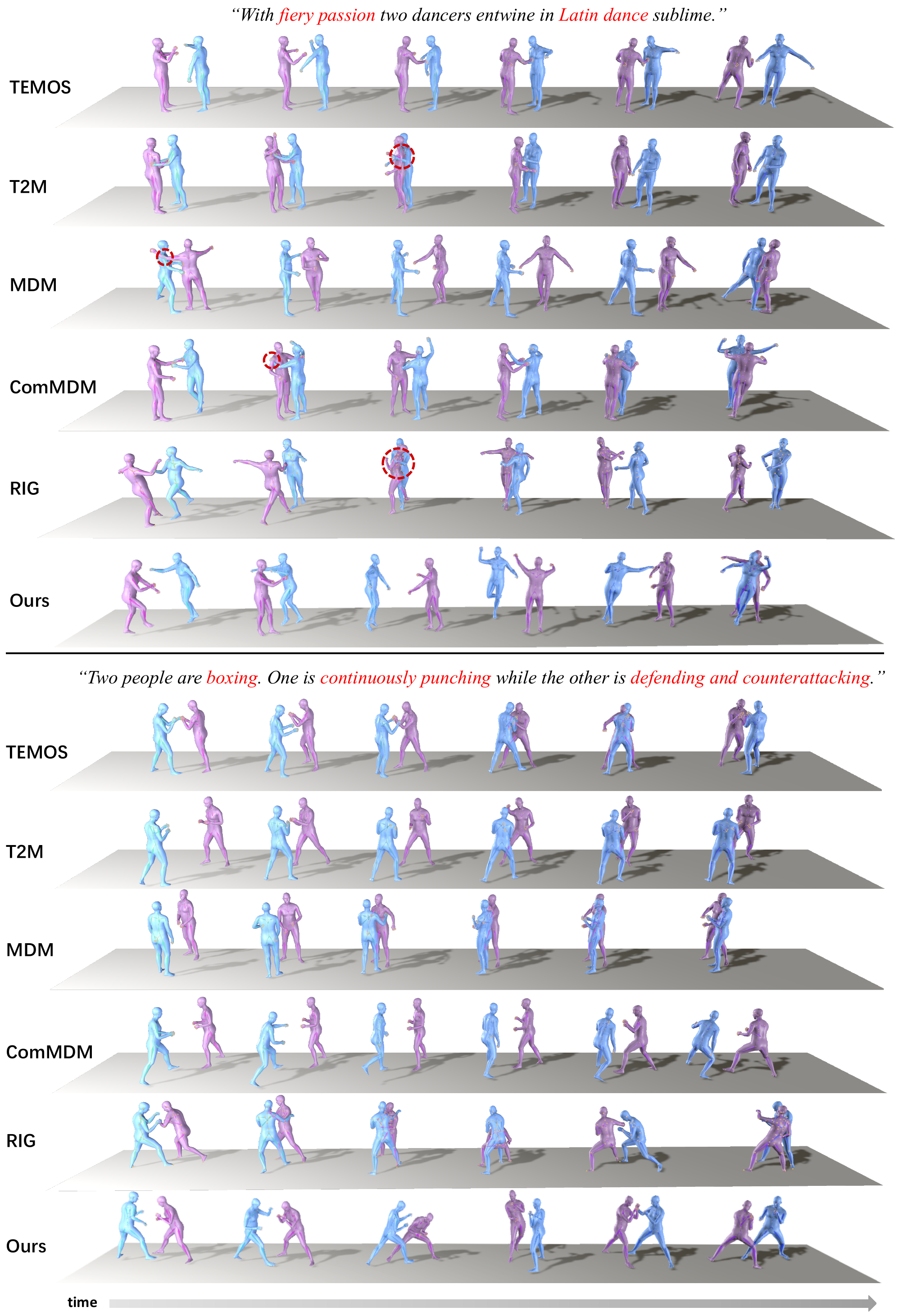} 
	\caption{\textbf{Qualitative comparison} with previous state-of-the-art works. The inputs to the model are listed at the top and middle, while the outputs of different models ~\citep{petrovich2022temos,guo2022generating,tevet2022human,shafir2023human} are listed below. Intersecting portions of the motions are highlighted with red dashed circles. } 
	
 \label{Fig.8} 
\end{figure*} 

\begin{table*}[htbp]
	\begin{center}
		\centering
		\caption{Quantitative evaluation of our key designs. The performance drop-off highlights our technical contributions.} \label{Tab.3}
		
 		\resizebox{1\textwidth}{!}{
		\begin{tabular}{lccccccc}
		    \toprule
		    \multirow{2}{*}{Methods}  & \multicolumn{3}{c}{R Precision$\uparrow$} & \multirow{2}{*}{FID $\downarrow$} & \multirow{2}{*}{MM Dist$\downarrow$}  & \multirow{2}{*}{Diversity$\rightarrow $} & \multirow{2}{*}{MModality $\uparrow$} \\
		    \cmidrule(lr){2-4}
			& Top 1 & Top 2  & Top 3 \\
			\midrule
                Real   &  $0.452^{\pm .008}$ &  $0.610^{\pm .009}$  & $0.701^{\pm .008}$   &  $0.273^{\pm .007}$ &  $3.755^{\pm .017}$  & $7.948^{\pm .008}$   & -  \\
                \midrule
            canonical rep     &  $0.134^{\pm .011}$ &  $0.233^{\pm .013}$  & $0.315^{\pm .013}$   &  $11.322^{\pm .055}$ &  $7.865^{\pm .015}$  & $8.534^{\pm .028}$   & $\textbf{2.963}^{\pm .068}$   \\
            concat-conditioning     &  $0.303^{\pm .010}$ &  $0.451^{\pm .010}$  & $0.575^{\pm .012}$   &  $6.273^{\pm .080}$ &  $5.579^{\pm .012}$  & $7.123^{\pm .035}$   & $1.790^{\pm .052}$   \\
			w/o weights sharing     &  $0.153^{\pm .015}$ &  $0.257^{\pm .013}$  & $0.337^{\pm .012}$   &  $8.059^{\pm .077}$ &  $7.242^{\pm .010}$  & $\textbf{7.508}^{\pm .026}$   & $2.288^{\pm .076}$   \\
            
			w/o DM loss       &  $0.293^{\pm .012}$ &  $0.437^{\pm .010}$  & $0.533^{\pm .016}$   &  $6.653^{\pm .064}$ &  $5.934^{\pm .011}$  & $7.328^{\pm .033}$   & $2.181^{\pm .055}$   \\
			w/o RO loss         &  $0.310^{\pm .013}$ &  $0.466^{\pm .009}$  & $0.587^{\pm .010}$   &  $6.311^{\pm .052}$ &  $5.515^{\pm .012}$  & $7.309^{\pm .050}$   & $2.019^{\pm .053}$   \\
			\midrule
            {InterGen (LLaMA-7B)}    &  {$0.345^{\pm .012}$} &  {$0.493^{\pm .012}$}  & {$0.605^{\pm .014}$}   &  {$\textbf{5.865}^{\pm .086}$} &  {$5.319^{\pm .017}$}  & {$7.459^{\pm .036}$} & {$2.212^{\pm .075}$} \\
			InterGen (CLIP-ViT-L/14)    &  $\textbf{0.371}^{\pm .010}$ &  $\textbf{0.515}^{\pm .012}$  & $\textbf{0.624}^{\pm .010}$   &  $5.918^{\pm .079}$ &  $\textbf{5.108}^{\pm .014}$  & $7.387^{\pm .029}$ & $2.141^{\pm .063}$ \\
			\bottomrule
		\end{tabular}
 		}
	\end{center}

\end{table*}

\begin{table*}[htbp]
	\begin{center}
		\centering
		\caption{Quantitative evaluation of our regularization loss schedule training scheme. The strategy of different treatments for different noise levels improves the performance significantly.} \label{Tab.4}
		
 		\resizebox{1\textwidth}{!}{
		\begin{tabular}{lccccccc}
		    \toprule
		    \multirow{2}{*}{RegLoss schedule}  & \multicolumn{3}{c}{R Precision$\uparrow$} & \multirow{2}{*}{FID $\downarrow$} & \multirow{2}{*}{MM Dist$\downarrow$}  & \multirow{2}{*}{Diversity$\rightarrow$} & \multirow{2}{*}{MModality $\uparrow$}  \\
		    \cmidrule(lr){2-4}
			& Top 1 & Top 2  & Top 3 \\
			\midrule
                Real motion        &  $0.452^{\pm .008}$ &  $0.610^{\pm .009}$  & $0.701^{\pm .008}$   &  $0.273^{\pm .066}$ &  $3.755^{\pm .015}$  & $7.948^{\pm .008}$   & -  \\
                \midrule
            None         &  $0.201^{\pm .010}$ &  $0.315^{\pm .014}$  & $0.406^{\pm .009}$   &  $7.862^{\pm .074}$ &  $6.919^{\pm .011}$  & $7.301^{\pm .044}$   & $\textbf{2.198}^{\pm .042}$  \\
			$ t \leq 0.1 T $        &  $0.285^{\pm .013}$ &  $0.443^{\pm .010}$  & $0.544^{\pm .014}$   &  $6.556^{\pm .065}$ &  $5.822^{\pm .013}$  & $7.315^{\pm .032}$   & $2.156^{\pm .055}$  \\
			$t\leq 0.2T   $        &  $0.310^{\pm .013}$ &  $0.464^{\pm .010}$  & $0.561^{\pm .014}$   &  $6.178^{\pm .086}$ &  $5.443^{\pm .012}$  & $7.342^{\pm .037}$   & $2.122^{\pm .078}$  \\
			$t\leq 0.3T  $         &  $0.338^{\pm .010}$ &  $0.483^{\pm .014}$  & $0.582^{\pm .011}$   &  $6.034^{\pm .069}$ &  $5.461^{\pm .009}$  & $7.323^{\pm .040}$   & $2.114^{\pm .039}$  \\
                $t\leq 0.4T $          &  $0.353^{\pm .010}$ &  $0.496^{\pm .014}$  & $0.598^{\pm .012}$   &  $5.945^{\pm .078}$ &  $5.263^{\pm .013}$  & $7.303^{\pm .035}$   & $2.095^{\pm .061}$  \\
                $t\leq 0.5T $    &  $0.362^{\pm .011}$ &  $0.506^{\pm .013}$  & $0.610^{\pm .014}$   &  $5.938^{\pm .052}$ &  $5.175^{\pm .016}$  & $7.318^{\pm .041}$   & $2.129^{\pm .046}$  \\
                $t\leq 0.6T $        &  $0.367^{\pm .013}$ &  $0.513^{\pm .009}$  & $0.619^{\pm .010}$   &  $\textbf{5.887}^{\pm .064}$ &  $5.209^{\pm .012}$  & $7.356^{\pm .022}$   & $2.133^{\pm .067}$ \\
                $t\leq 0.7T $    &  $\textbf{0.371}^{\pm .010}$ &  $\textbf{0.515}^{\pm .012}$  & $\textbf{0.624}^{\pm .010}$   &  $5.918^{\pm .079}$ &  $\textbf{5.108}^{\pm .014}$  & $\textbf{7.387}^{\pm .029}$ & $2.141^{\pm .063}$ \\
               $t \leq 0.8T$        &  $0.299^{\pm .012}$ &  $0.453^{\pm .011}$  & $0.555^{\pm .014}$   &  $6.521^{\pm .070}$ &  $5.745^{\pm .012}$  & $7.346^{\pm .024}$   & $2.177^{\pm .062}$  \\
                $t\leq  0.9T $       &  $0.278^{\pm .010}$ &  $0.432^{\pm .009}$  & $0.532^{\pm .013}$   &  $6.664^{\pm .062}$ &  $6.042^{\pm .012}$  & $7.314^{\pm .039}$   & $2.103^{\pm .072}$  \\
                $t\leq  T  $     &  $0.232^{\pm .012}$ &  $0.365^{\pm .011}$  & $0.468^{\pm .010}$   &  $7.037^{\pm .053}$ &  $6.620^{\pm .010}$  & $7.282^{\pm .028}$   & $2.135^{\pm .080}$ \\

			\bottomrule
		\end{tabular}
 		}
	\end{center}

\end{table*}

\section{Experiments}
\label{experimental results}
Here, we demonstrate the capability of our approach in a variety of scenarios. We first introduce the evaluation dataset and metrics and then showcase a gallery of our generation results for two-person interactions in Fig.~\ref{Fig.7}. We then provide the comparison with previous methods as well as the evaluation of our technical components, both qualitatively and quantitatively, followed by the analysis of various downstream applications. 

\myparagraph{Evaluation dataset.}\label{evaluation dataset}
The existing available human motion datasets lack sufficient categories of human interactions and corresponding text descriptions. We hence contribute a new dataset, InterHuman, to evaluate our approach. It provides accurate skeletal human motions and rich natural language descriptions, covering diverse two-person interaction scenarios (see Sec.~\ref{interhuman dataset} for more details).  We also augment our data following the HumanML3D~\citep{guo2022generating}, which involves mirroring all motions and replacing relevant keywords in the descriptions, and swapping the order of two people in all interactions. We then split the data into training, validation, and test sets using the same protocol.

\myparagraph{Evaluation metrics.}\label{evaluation metrics}
{
Following single-person text-to-motion generation, we adopt the same evaluation metrics as~\cite{guo2022generating}, which are listed as follows:
\begin{enumerate}
    \item \emph{R-Precision}. To measure the text-motion consistency, we rank the Euclidean distances between the motion and text embeddings. Top-1, Top-2, and Top-3 accuracy of motion-to-text retrieval are reported.
    \item \emph{Frechet Inception Distance (FID)}. To measure the similarity between synthesized and real interactive motions, we calculate the latent embedding distribution distance between the generated and real interactive motions using FID~\citep{heusel2017gans} on the extracted motion features.
    \item \emph{Multimodal Distance (MM Dist)}. To measure the similarity between each text and the corresponding motion, the average Euclidean distance between each text embedding and the
generated motion embedding from this text is reported.
    \item \emph{Diversity}. We randomly sample 300 pairs of motions and calculate the average Euclidean distances of the pairs in latent space to measure motion diversity in the generated motion dataset.
    \item \emph{Multimodality (MModality)}. Similar to \emph{Diversity}, we sample 20 motions within one text prompt to form 10 pairs, and measure the average latent Euclidean distances of the pairs. The average over all the text descriptions is reported.
\end{enumerate}
}
To calculate these metrics, we train an interaction motion feature extractor and a text feature extractor using contrastive loss following~\cite{radford2021learning}, which encourages matched text-interaction pairs to have geometrically close feature vectors.

\begin{figure*}[htbp] 
	\centering  
	\includegraphics[width=1.0\linewidth]{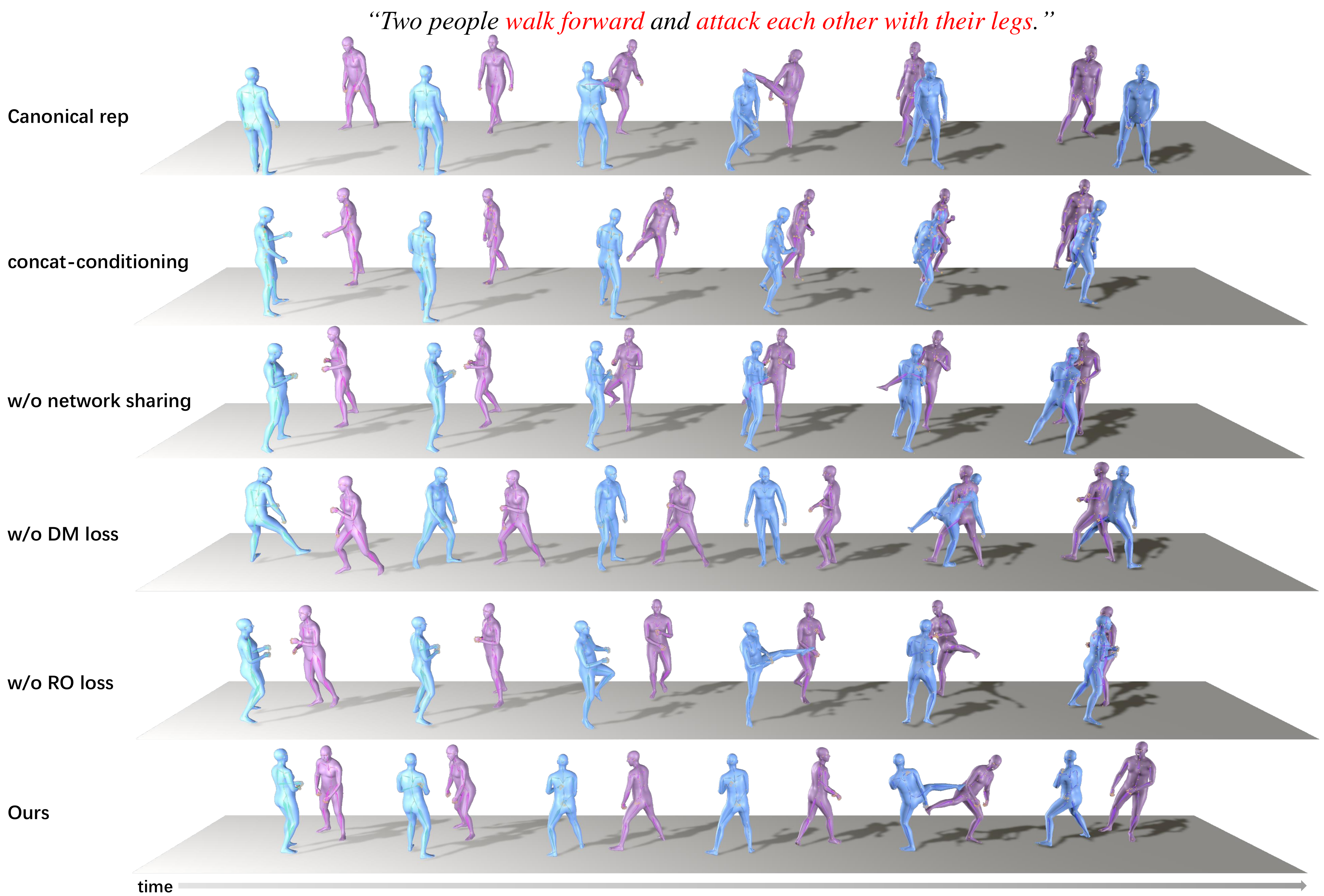} 
	\caption{Qualitative results of ablation study. The top of the figure displays text prompts, while the lower illustrates the results of different ablation experiments and our best result. For quantitative comparisons of experimental results, please refer to Tab.~\ref{Tab.3}.} 
 \label{Fig.9} 
\end{figure*}

\begin{figure*}[htb] 
	\centering  
	\includegraphics[width=1.0\linewidth]{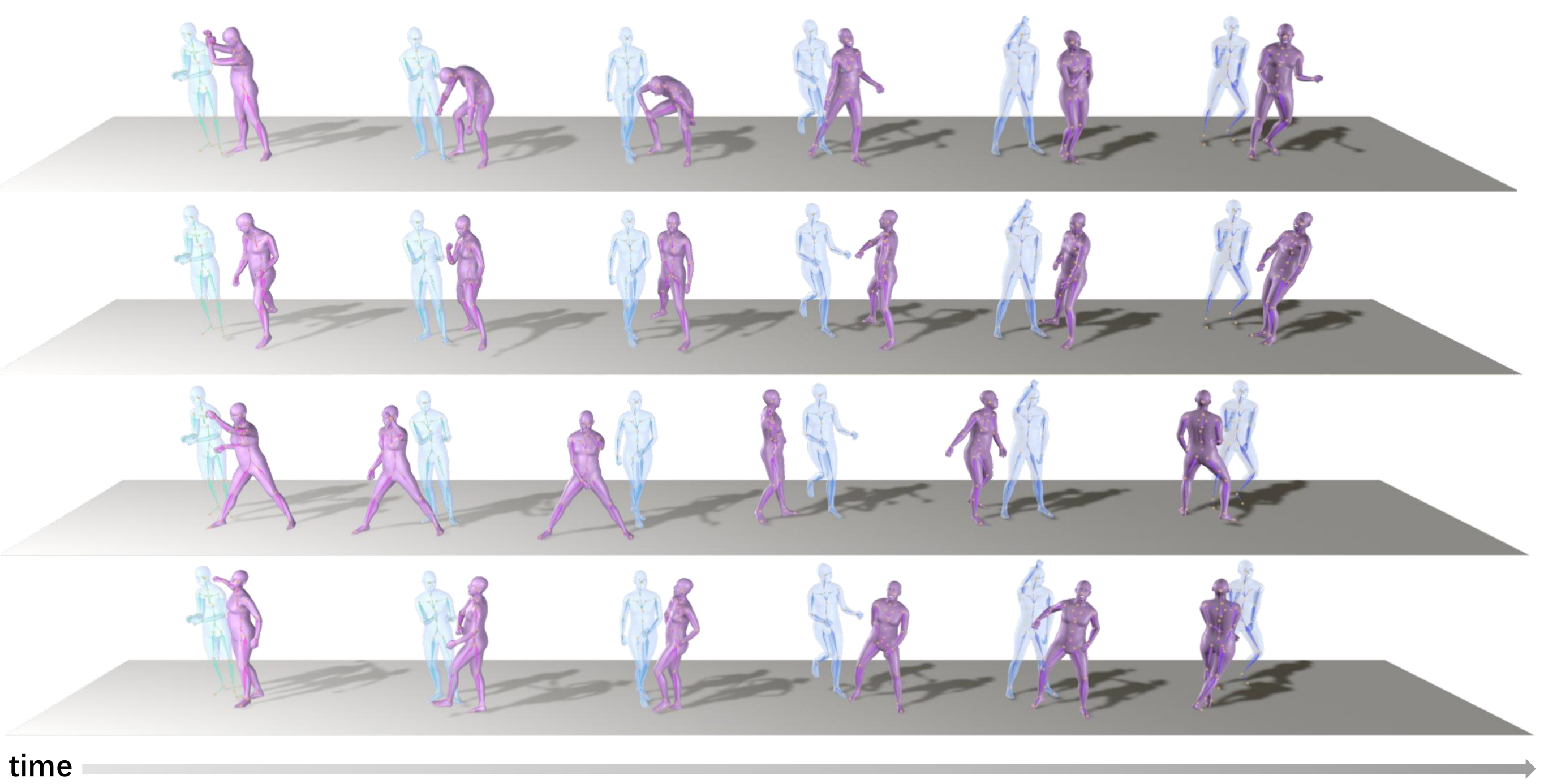} 
	\caption{\textbf{Person-to-person generation.} The above five motions are generated with the premise of freezing the motion of one person (represented by a semi-transparent SMPL model) while generating the motion of the other person (represented by an opaque 
SMPL model).} 
	 
 \label{Fig.10}
\end{figure*}

\subsection{Implementation Details}\label{implementation details}

We implement our InterGen with $N=8$ blocks whose latent dimension is set to 1024 and each attention layer consists of 8 heads, as same as the re-implemented MDM and ComMDM.
We employ a frozen \emph{CLIP-ViT-L/14} model as the text encoder.
The number of diffusion timesteps is set to 1,000 during training and we apply the DDIM~\citep{song2020denoising} sampling strategy with 50 timesteps and $\eta = 0$. 
We adopt the cosine noise level schedule~\citep{nichol2021improved} and classifier-free guidance~\citep{ho2022classifier} where the $10\%$ random CLIP embeddings are set to zero during training and the guidance coefficient is set to 3.5 during sampling.
{All the models are trained with AdamW~\citep{loshchilov2017decoupled} optimizer with betas of (0.9, 0.999), a weight decay of $2\times10^{-5}$, a maximum learning rate of $10^{-4}$, and a cosine LR schedule with 10 linear warm-up epochs. }
{To address the disparity among several loss terms in magnitude orders present in Eqn.~\ref{eqn:regloss}, we reweight them using multiple hyper-parameters, which harmonize these terms into a similar magnitude order, thereby ensuring a balanced contribution from each term.
Specifically, we set $\lambda_{vel}=30$, $\lambda_{foot}=30$, $\lambda_{BL}=10$, $\lambda_{DM}=3$, and $\lambda_{RO}=0.01$. }
For Eqn.~\ref{eqn:loss}, we set $\lambda_{reg}=1$ in all the experiments.
We train our diffusion denoisers with a batch size of 64 for 2,000 epochs on two Nvidia Tesla V100 GPUs, 
{which takes over 70 hours.
We run the inference on a PC with an Intel Core i7-10700K CPU and an Nvidia RTX 3080ti GPU, which takes an average of 8s to generate a motion sequence of 300 frames with a frame rate of 30.}

\subsection{Comparisons}\label{comparison}
We compare our InterGen with various representative text-to-motion methods in two-person interactive scenarios.
Specifically, we apply single-person methods VAE-based TEMOS~\citep{petrovich2022temos} and T2M~\citep{guo2022generating}, diffusion-based MDM~\citep{tevet2022human}, and recent two-person method ComMDM~\citep{shafir2023human} and RIG~\citep{tanaka2023role}.
To thoroughly evaluate our method and conduct fair comparisons, we retrain the above methods with the same InterHuman training set and test on the test set.
Note that for extending the above single-person motion synthesis models to handle two-person interaction, we modify the input and output dimensions of their networks to accommodate our non-canonical representation of two-person interaction.
For fair comparisons, we report the results of ComMDM pre-trained and fine-tuned in the original few-shot setting with 10 training samples (with *) and retrained on the same InterHuman training set (without *) with the same data representation. Note that the source code and training data of ComMDM are not publicly available yet and we re-implement it with the same setting on our dataset. 

Tab.~\ref{Tab.2} summarizes the quantitative comparison results. Specifically, our approach outperforms other baselines in terms of FID, R precision, and MM Dist. Note that these metrics numbers are calculated using the whole test sets, which indicates that our InterGen achieves more compelling interaction motion generation with more accurate text/motion matching. 
The corresponding representative qualitative results are provided in Fig.~\ref{Fig.8}, which demonstrates that our approach ensures the diversity and plausibility of the generated two-person interaction motions. Note that our approach can generate more natural interaction states, more diverse motions, and more accurate alignment of the global relative orientations and translations between the two performers.

\subsection{Evaluations} \label{ablation studies}
Here, we provide ablation analysis for the key technical designs in our InterGen approach.

\myparagraph{Interaction motion representation.}\label{evaluation on interaction representation}
We first compare our non-canonical representation with the canonical representation, as shown in $canonical ~rep$ row in Tab.~\ref{Tab.3}. 
We replace the motion representation using canonical representation and retrain our model in the same setting. The R precision and FID drop off significantly, which indicates the lower quality of motions and text prompt matching. The qualitative result is shown in the first row of Fig.~\ref{Fig.9}. 
{As time flows, the cumulative integration errors of the trajectories exponentially expand, which leads to two people performing their own motions without any spatial relation to each other, making interactions become unrealistic.
This demonstrates the superiority of our two-person non-canonical motion representation in the common frame, which explicitly preserves the spatial relations of two people in the same space. 
It avoids the harmful cumulative integration process and facilitates the model to directly learn the spatial relations in the same frame without complex frame transformations.}

\begin{figure*}[tbp] 
	\centering  
	\includegraphics[width=1.0\linewidth]{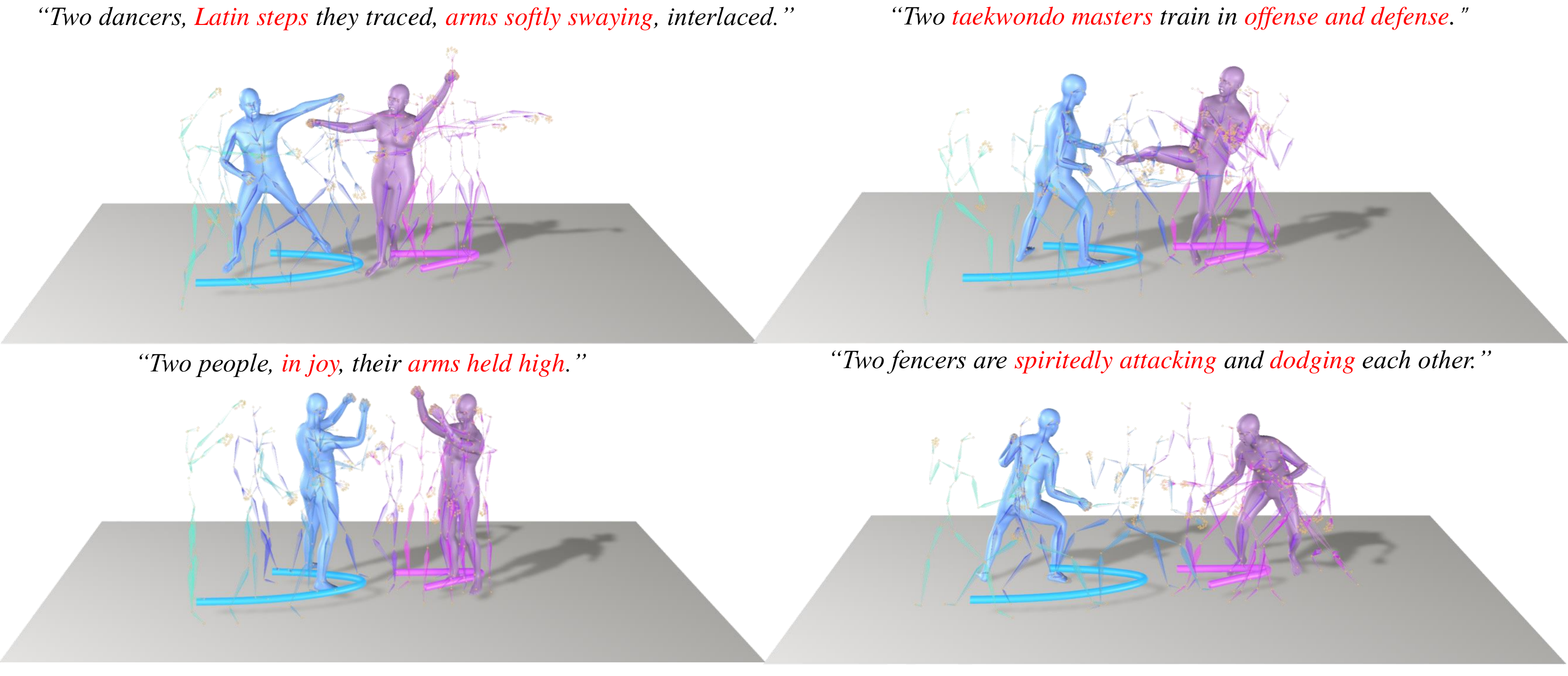} 
	\caption{\textbf{Trajectory control.} The curve beneath the peoples' feet represents their motion trajectories. The skeleton representation displays the position and motion of each frame over time, with the SMPL model~\citep{loper2015smpl} indicating a specific frame of motion. The text input for each motion is provided above the respective persons.} 
	
 \label{Fig.11} 
\end{figure*}

\myparagraph{Cooperative networks.}\label{evaluation on cooperative networks}
Next, we evaluate the effectiveness of our cooperative networks. 
We first replace the mutual attention conditioning mechanism with concatenation, where each network concatenates its own noisy motion with the counterpart noisy motion as the input and drop the mutual attention layers in its blocks and retrain the model.
The quantitative result is shown in the $concat-conditioning$ row in Tab.~\ref{Tab.3}, which shows a decline in performance. The qualitative result is shown in the second row of Fig.~\ref{Fig.9}, where the plausibility of the interaction and alignment of the motions are significantly reduced compared to ours.

We then ablate the weights sharing mechanism and train two networks without explicitly enforcing them to be the same. The quantitative result is presented in the $w/o\_weights\_sharing$ row in Tab.~\ref{Tab.3}, where R precision and FID are significantly worse than ours, as also illustrated in the third row of Fig.~\ref{Fig.9}, where two people do not show similar activation and motion capacity during the interaction. 
These indicate that our cooperative networks with mutual attention and weights sharing mechanisms can effectively handle not only the complexity of interaction but also the balance of motion capacity between two people.

\myparagraph{Interaction losses.}\label{evaluation on interactive losses}
We also apply ablation experiments on our two interactive losses and a regularization loss schedule.
As shown in the $w/o\_DM\_loss$ row of Tab.~\ref{Tab.3}, we drop the DM loss and retrain our model, the interaction generation performance drops off, as shown in the fourth row of Fig.~\ref{Fig.9} qualitatively, the two people perform a weird interaction that they pass through each other's body since the absence of distance constraints.

And we drop the RO loss and apply the same process, as shown in the $w/o\_RO\_loss$ row of Tab.~\ref{Tab.3}, getting a worse performance too. The qualitative result in the fifth row of Fig.~\ref{Fig.9} demonstrates that the two people perform an unrealistic interaction where the relative orientation between them does not match the motions they perform.
These qualitative and quantitative results demonstrate the effectiveness and necessity of our interactive losses.

Tab.~\ref{Tab.4} demonstrates the effect of the choice of diffusion timesteps to apply regularization loss. 
We choose the model with the best performance occurring on $t \leq 0.7 T$ with a cosine diffusion noise level schedule~\citep{nichol2021improved} as our final model, which outperforms the naively applying to all timesteps ($t \leq T$) and not applying such a regularization loss ($None$).
This indicates the effectiveness of our loss schedule training scheme.

\myparagraph{Pre-trained text encoders.}\label{evaluation on text encoder}
{We additionally explore the effect of pre-trained text encoders for the human interaction generation task.
We choose the widely employed CLIP~\citep{radford2021learning} model in motion generation~\citep{tevet2022motionclip,tevet2022human,shafir2023human,zhang2023t2m} and recent open-source large language model LLaMA~\citep{touvron2023llama} equipped with the powerful generative pre-training technique~\citep{brown2020language}.
Specifically, we replace the \emph{CLIP-ViT-L/14} text encoder with the \emph{LLaMA-7B} language model, where we also freeze its parameters and extract the text embeddings from its penultimate layer.
The results are shown in Tab.~\ref{Tab.3}, we observe that simply replacing the text encoder with LLM leads to worse R-precision and MM Dist that indicate a damaged text-motion consistency, although it improves the FID slightly.}

\begin{figure*}[tbp] 
	\centering  
	\includegraphics[width=1.0\linewidth]{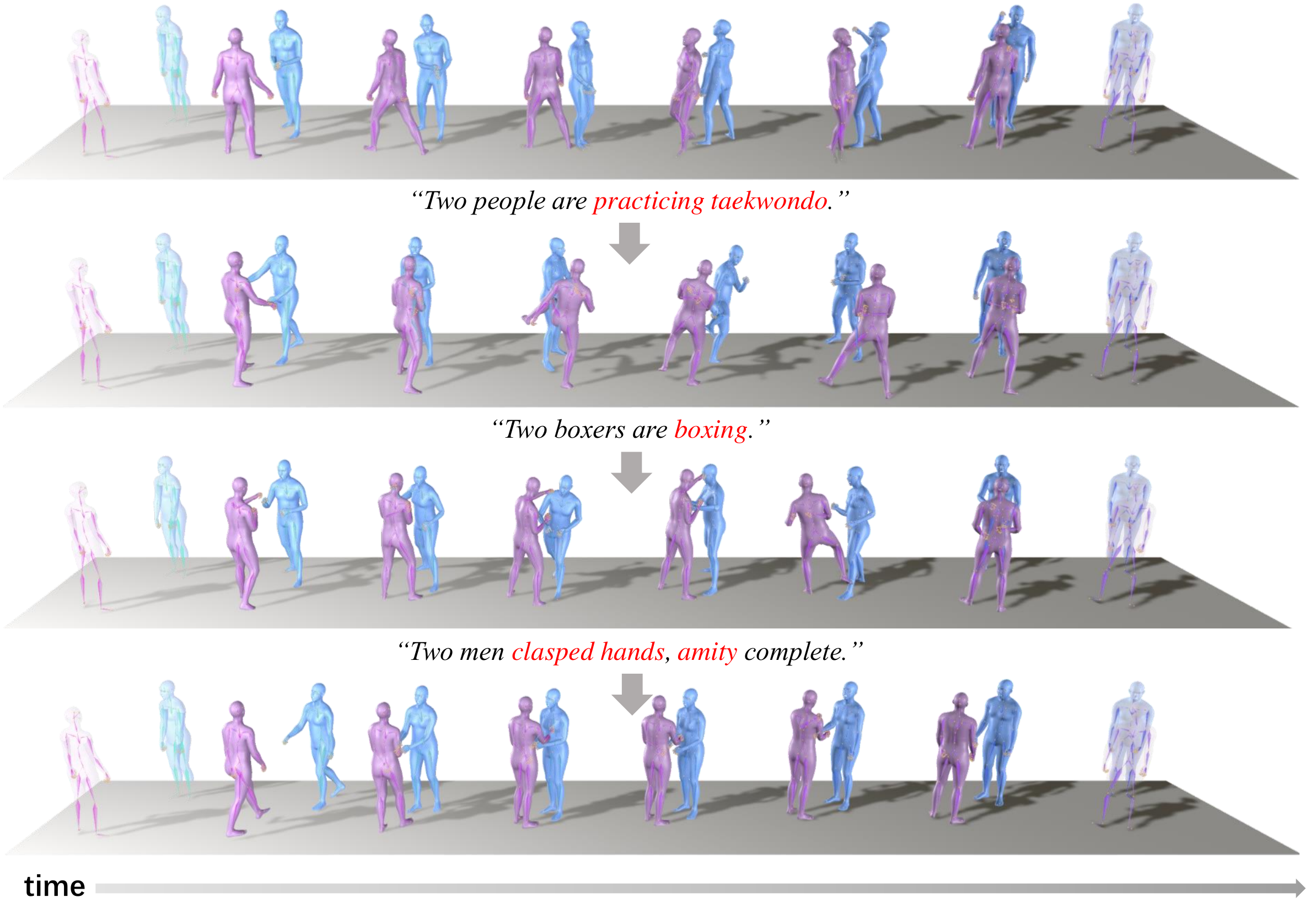} 
	\caption{\textbf{Interaction inbetweening}. The first row depicts the outcome of motion freezing the starting and ending motions, without the use of any text prompts. Subsequently, the output generated in the second row is used as the input for text prompts, resulting in the in-betweening results presented in the next row. By iterating this process, the motion sequence is gradually transformed into a daily motion, as shown in the final row.} 

 \label{Fig.12} 
\end{figure*}

\subsection{Applications}\label{application}
Once trained, our InterGen model can be easily customized and employed in various human interaction-related tasks. Here, we showcase a series of downstream applications using our InterGen, i.e., human-to-human motion generation, additional trajectory control on top of text-prompt, and generating diverse motion inbetweening for interaction scenes.

\myparagraph{Person-to-person generation.}\label{person-to-person generation}
Here we fine-tune the original InterGen to enable the generation of human-to-human motions by simply taking a single-person motion sequence as input. Our method works by masking the noise applied to the motions that we wish to freeze during the forward pass of the diffusion process. The frozen motion serves as ground truth that propagates into the model and hence the model learns to rely on these motions when attempting to reconstruct the counterpart motions. This workflow is inspired by the fine-tuning strategy from MDM~\citep{tevet2022human} and please refer to the original paper for implementation details. Differently, we freeze one person's motion instead of setting the trajectory to zero and utilize two inputs $\{\textbf{x}_a$,$\textbf{x}_b\}$ rather than one input $\textbf{x}_0$. The visualization of this person-to-person generation process is shown in Fig.~\ref{Fig.10}. The translucent person represents the frozen motion, i.e., the 
ground-truth motion at each iteration, while the opaque person represents the conditionally generated motions. It can be observed that our diffusion process produces diverse motion results.

\myparagraph{Trajectory control.}\label{trajectory control}
We further demonstrate combining additional trajectory controls with the text inputs for more controllable human-to-human interaction generation. Specifically, it requires pre-inputting the trajectory during the text processing stage to mask the global transform of motion during the diffusion process. Such a strategy allows us to apply the trajectory control injected in the input stage to the diffusion at each time step, thus easily obtaining the interaction motions of two people that conform to the trajectory. 
As shown in Fig.~\ref{Fig.11}, the trajectory of the human-to-human interaction and the differently stylized interaction actions under this trajectory are clearly depicted. The skeleton in the figure represents the historical trajectory of the motion, and the two visualized SMPL models~\citep{loper2015smpl} represent the end of the motion. Under the constraint of the same trajectory, our approach can generate diverse and naturally conformed motions.

\myparagraph{Interaction inbetweening.}\label{interaction inbetweening}
Similar to person-to-person generation, we can also fine-tune the InterGen to freeze the beginning and ending segments of the motion sequence, so as to enforce InterGen to generate diverse Interaction motions that fill the frozen segments in between for motion inbetweening applications. 
Specifically, we implement time freezing of human-to-human motion in the pipeline and freeze motions instead of the training step during sampling. We then replace the part of in-between motion sequence inputs at each timestep when sampling. As shown in Fig.~\ref{Fig.12}, the translucent double models represent the motion that has been frozen at the initial and end time steps, and the opaque double models represent the generated motions. Note that our approach can generate vivid motion in-betweening results that are also incorporated with the text-prompt inputs, even under various interaction scenarios.

\subsection{Limitation and Discussion}\label{limitation}
Our InterGen has achieved compelling results for generating realistic two-person interaction motions from text prompts, yet still yielding some limitations. 
First, our approach only considers the interactions between two people, which may limit the applicability of our approach to more complex scenarios (e.g., crowd dynamics simulation). It's promising to model more complex group interactions such as team sports or group dances, yet it requires much more diverse training datasets. 
Besides, our approach only generates motion sequences based on given text prompts, without further considering the feedback from the user. This may limit the user’s creativity over the generated motions, especially when the text prompts are vague or ambiguous. It's an interesting direction to generate motions that satisfy the user’s specific preferences or expectations, such as the style, speed, or intensity of the motions. 
Our generated motion is also limited to a fixed largest length, which cannot support extremely long-sequence generation. It hence hinders the diversity and coherence of the generated motions, especially when the text prompts are complex. More advanced strategies like involving multiple sub-motions or transitions to span a longer period of time will help.
{Moreover, our generated motions still suffer from jittering artifact, which is a known challenge in motion generation especially when maintaining a consistent trajectory over longer sequences. A more carefully designed temporal smoothing can further alleviate it, yet may sacrifice the quality and diversity of the generated motions. 
Besides, penetration artifacts sometimes occur, which probably can be improved by adding more self-contact constraints and further introducing physics simulation. 
We leave these issues to future exploration where our dataset and approach can serve as the cornerstone and a strong baseline.}

\section{Conclusion}

{In this paper, we present InterGen,} a diffusion-based approach to conveniently generate two-person motion under diverse interactions, from only text-prompt controls.
Specifically, we contribute a novel multimodal dataset with rich motion results and natural language descriptions, covering a wide range of interaction scenarios.
Then, in our interaction diffusion model, our cooperative denoisers with sharing weights and a mutual attention mechanism can effectively model the symmetric fact of human identities during interactions.
Our non-canonical motion representation also effectively models the global relations between performers for the interaction setting.
Our regularization design with a specific damping scheme further encodes the spatial relations to generate more diverse and reasonable interactions. 
Extensive experimental results have demonstrated the effectiveness of
InterGen for the generation of compelling two-person motions and a series of downstream interaction applications. 
We believe our approach and multimodal dataset can serve as a solid step towards text-guided generation and understanding of human-to-human interactions, with numerous potential applications for entertainment, gaming, and immersive experience in VR/AR.


{\small
\bibliographystyle{spbasic}
\bibliography{egbib}

\begin{thebibliography}{103}
\providecommand{\natexlab}[1]{#1}
\providecommand{\url}[1]{{#1}}
\providecommand{\urlprefix}{URL }
\expandafter\ifx\csname urlstyle\endcsname\relax
  \providecommand{\doi}[1]{DOI~\discretionary{}{}{}#1}\else
  \providecommand{\doi}{DOI~\discretionary{}{}{}\begingroup \urlstyle{rm}\Url}\fi
\providecommand{\eprint}[2][]{\url{#2}}

\bibitem[{Van~der Aa et~al.(2011)Van~der Aa, Luo, Giezeman, Tan, and Veltkamp}]{van2011umpm}
Van~der Aa N, Luo X, Giezeman GJ, Tan RT, Veltkamp RC (2011) Umpm benchmark: A multi-person dataset with synchronized video and motion capture data for evaluation of articulated human motion and interaction. In: 2011 IEEE international conference on computer vision workshops (ICCV Workshops), IEEE, pp 1264--1269

\bibitem[{Ahn et~al.(2018)Ahn, Ha, Choi, Yoo, and Oh}]{ahn2018text2action}
Ahn H, Ha T, Choi Y, Yoo H, Oh S (2018) Text2action: Generative adversarial synthesis from language to action. In: 2018 IEEE International Conference on Robotics and Automation (ICRA), IEEE, pp 5915--5920

\bibitem[{Ahuja and Morency(2019)}]{ahuja2019language2pose}
Ahuja C, Morency LP (2019) Language2pose: Natural language grounded pose forecasting. In: 2019 International Conference on 3D Vision (3DV), IEEE, pp 719--728

\bibitem[{Andrews et~al.(2016)Andrews, Huerta, Komura, Sigal, and Mitchell}]{Andrews2016}
Andrews S, Huerta I, Komura T, Sigal L, Mitchell K (2016) Real-time physics-based motion capture with sparse sensors. In: Proceedings of the 13th European conference on visual media production (CVMP 2016), pp 1--10

\bibitem[{Anguelov et~al.(2005)Anguelov, Srinivasan, Koller, Thrun, Rodgers, and Davis}]{anguelov2005scape}
Anguelov D, Srinivasan P, Koller D, Thrun S, Rodgers J, Davis J (2005) Scape: shape completion and animation of people. In: ACM SIGGRAPH 2005 Papers, pp 408--416

\bibitem[{Ao et~al.(2022)Ao, Gao, Lou, Chen, and Liu}]{ao2022rhythmic}
Ao T, Gao Q, Lou Y, Chen B, Liu L (2022) Rhythmic gesticulator: Rhythm-aware co-speech gesture synthesis with hierarchical neural embeddings. ACM Transactions on Graphics (TOG) 41(6):1--19

\bibitem[{Athanasiou et~al.(2022)Athanasiou, Petrovich, Black, and Varol}]{athanasiou2022teach}
Athanasiou N, Petrovich M, Black MJ, Varol G (2022) Teach: Temporal action composition for 3d humans. In: 2022 International Conference on 3D Vision (3DV), IEEE, pp 414--423

\bibitem[{Bogo et~al.(2016)Bogo, Kanazawa, Lassner, Gehler, Romero, and Black}]{bogo2016keep}
Bogo F, Kanazawa A, Lassner C, Gehler P, Romero J, Black MJ (2016) Keep it smpl: Automatic estimation of 3d human pose and shape from a single image. In: Computer Vision--ECCV 2016: 14th European Conference, Amsterdam, The Netherlands, October 11-14, 2016, Proceedings, Part V 14, Springer, pp 561--578

\bibitem[{Bregler and Malik(1998)}]{bregler1998tracking}
Bregler C, Malik J (1998) Tracking people with twists and exponential maps. In: Proceedings. 1998 IEEE Computer Society Conference on Computer Vision and Pattern Recognition (Cat. No. 98CB36231), IEEE, pp 8--15

\bibitem[{Brown et~al.(2020)Brown, Mann, Ryder, Subbiah, Kaplan, Dhariwal, Neelakantan, Shyam, Sastry, Askell et~al.}]{brown2020language}
Brown T, Mann B, Ryder N, Subbiah M, Kaplan JD, Dhariwal P, Neelakantan A, Shyam P, Sastry G, Askell A, et~al. (2020) Language models are few-shot learners. Advances in neural information processing systems 33:1877--1901

\bibitem[{Chen et~al.(2022)Chen, Su, Yang, Cheng, Xu, Fu, and Yu}]{chen2022learning}
Chen X, Su Z, Yang L, Cheng P, Xu L, Fu B, Yu G (2022) Learning variational motion prior for video-based motion capture. arXiv preprint arXiv:221015134

\bibitem[{Chen et~al.(2023)Chen, Jiang, Liu, Huang, Fu, Chen, and Yu}]{chen2022executing}
Chen X, Jiang B, Liu W, Huang Z, Fu B, Chen T, Yu G (2023) Executing your commands via motion diffusion in latent space. In: Proceedings of the IEEE/CVF Conference on Computer Vision and Pattern Recognition, pp 18000--18010

\bibitem[{De~Aguiar et~al.(2008)De~Aguiar, Stoll, Theobalt, Ahmed, Seidel, and Thrun}]{de2008performance}
De~Aguiar E, Stoll C, Theobalt C, Ahmed N, Seidel HP, Thrun S (2008) Performance capture from sparse multi-view video. In: ACM SIGGRAPH 2008 papers, pp 1--10

\bibitem[{Duan et~al.(2021)Duan, Shi, Zou, Lin, Qian, Zhang, and Yuan}]{duan2021single}
Duan Y, Shi T, Zou Z, Lin Y, Qian Z, Zhang B, Yuan Y (2021) Single-shot motion completion with transformer. arXiv preprint arXiv:210300776

\bibitem[{Gall et~al.(2010)Gall, Rosenhahn, Brox, and Seidel}]{Gall2010}
Gall J, Rosenhahn B, Brox T, Seidel HP (2010) Optimization and filtering for human motion capture. International Journal of Computer Vision (IJCV) 87(1--2):75--92

\bibitem[{Ghosh et~al.(2021)Ghosh, Cheema, Oguz, Theobalt, and Slusallek}]{ghosh2021synthesis}
Ghosh A, Cheema N, Oguz C, Theobalt C, Slusallek P (2021) Synthesis of compositional animations from textual descriptions. In: Proceedings of the IEEE/CVF international conference on computer vision, pp 1396--1406

\bibitem[{Gilbert et~al.(2019)Gilbert, Trumble, Malleson, Hilton, and Collomosse}]{gilbert2019fusing}
Gilbert A, Trumble M, Malleson C, Hilton A, Collomosse J (2019) Fusing visual and inertial sensors with semantics for 3d human pose estimation. International Journal of Computer Vision 127:381--397

\bibitem[{Goodfellow et~al.(2020)Goodfellow, Pouget-Abadie, Mirza, Xu, Warde-Farley, Ozair, Courville, and Bengio}]{goodfellow2020generative}
Goodfellow I, Pouget-Abadie J, Mirza M, Xu B, Warde-Farley D, Ozair S, Courville A, Bengio Y (2020) Generative adversarial networks. Communications of the ACM 63(11):139--144

\bibitem[{Guo et~al.(2020)Guo, Zuo, Wang, Zou, Sun, Deng, Gong, and Cheng}]{guo2020action2motion}
Guo C, Zuo X, Wang S, Zou S, Sun Q, Deng A, Gong M, Cheng L (2020) Action2motion: Conditioned generation of 3d human motions. In: Proceedings of the 28th ACM International Conference on Multimedia, pp 2021--2029

\bibitem[{Guo et~al.(2022{\natexlab{a}})Guo, Zou, Zuo, Wang, Ji, Li, and Cheng}]{guo2022generating}
Guo C, Zou S, Zuo X, Wang S, Ji W, Li X, Cheng L (2022{\natexlab{a}}) Generating diverse and natural 3d human motions from text. In: Proceedings of the IEEE/CVF Conference on Computer Vision and Pattern Recognition, pp 5152--5161

\bibitem[{Guo et~al.(2022{\natexlab{b}})Guo, Zuo, Wang, and Cheng}]{guo2022tm2t}
Guo C, Zuo X, Wang S, Cheng L (2022{\natexlab{b}}) Tm2t: Stochastic and tokenized modeling for the reciprocal generation of 3d human motions and texts. In: Computer Vision--ECCV 2022: 17th European Conference, Tel Aviv, Israel, October 23--27, 2022, Proceedings, Part XXXV, Springer, pp 580--597

\bibitem[{Guo et~al.(2022{\natexlab{c}})Guo, Bie, Alameda-Pineda, and Moreno-Noguer}]{guo2022multi}
Guo W, Bie X, Alameda-Pineda X, Moreno-Noguer F (2022{\natexlab{c}}) Multi-person extreme motion prediction. In: Proceedings of the IEEE/CVF Conference on Computer Vision and Pattern Recognition, pp 13053--13064

\bibitem[{Habermann et~al.(2019)Habermann, Xu, Zollh\"{o}fer, Pons-Moll, and Theobalt}]{LiveCap2019tog}
Habermann M, Xu W, Zollh\"{o}fer M, Pons-Moll G, Theobalt C (2019) Livecap: Real-time human performance capture from monocular video. ACM Transactions on Graphics (TOG) 38(2):14:1--14:17

\bibitem[{Habermann et~al.(2020)Habermann, Xu, Zollhofer, Pons-Moll, and Theobalt}]{DeepCap_CVPR2020}
Habermann M, Xu W, Zollhofer M, Pons-Moll G, Theobalt C (2020) Deepcap: Monocular human performance capture using weak supervision. In: Proceedings of the IEEE/CVF Conference on Computer Vision and Pattern Recognition (CVPR)

\bibitem[{Habibie et~al.(2022)Habibie, Elgharib, Sarkar, Abdullah, Nyatsanga, Neff, and Theobalt}]{habibie2022motion}
Habibie I, Elgharib M, Sarkar K, Abdullah A, Nyatsanga S, Neff M, Theobalt C (2022) A motion matching-based framework for controllable gesture synthesis from speech. In: ACM SIGGRAPH 2022 Conference Proceedings, pp 1--9

\bibitem[{Harvey et~al.(2020)Harvey, Yurick, Nowrouzezahrai, and Pal}]{harvey2020robust}
Harvey FG, Yurick M, Nowrouzezahrai D, Pal C (2020) Robust motion in-betweening. ACM Transactions on Graphics (TOG) 39(4):60--1

\bibitem[{He et~al.(2021)He, Pang, Chen, Liang, Wu, Ma, and Xu}]{challencap}
He Y, Pang A, Chen X, Liang H, Wu M, Ma Y, Xu L (2021) Challencap: Monocular 3d capture of challenging human performances using multi-modal references. In: Proceedings of the IEEE/CVF Conference on Computer Vision and Pattern Recognition, pp 11400--11411

\bibitem[{Helten et~al.(2013)Helten, Muller, Seidel, and Theobalt}]{helten2013real}
Helten T, Muller M, Seidel HP, Theobalt C (2013) Real-time body tracking with one depth camera and inertial sensors. In: Proceedings of the IEEE international conference on computer vision, pp 1105--1112

\bibitem[{Henschel et~al.(2020)Henschel, Von~Marcard, and Rosenhahn}]{henschel2020accurate}
Henschel R, Von~Marcard T, Rosenhahn B (2020) Accurate long-term multiple people tracking using video and body-worn imus. IEEE Transactions on Image Processing 29:8476--8489

\bibitem[{Heusel et~al.(2017)Heusel, Ramsauer, Unterthiner, Nessler, and Hochreiter}]{heusel2017gans}
Heusel M, Ramsauer H, Unterthiner T, Nessler B, Hochreiter S (2017) Gans trained by a two time-scale update rule converge to a local nash equilibrium. Advances in neural information processing systems 30

\bibitem[{Ho and Salimans(2021)}]{ho2022classifier}
Ho J, Salimans T (2021) Classifier-free diffusion guidance. In: NeurIPS 2021 Workshop on Deep Generative Models and Downstream Applications

\bibitem[{Ho et~al.(2020)Ho, Jain, and Abbeel}]{ho2020denoising}
Ho J, Jain A, Abbeel P (2020) Denoising diffusion probabilistic models. Advances in Neural Information Processing Systems 33:6840--6851

\bibitem[{Huang et~al.(2017)Huang, Bogo, Lassner, Kanazawa, Gehler, Romero, Akhter, and Black}]{huang2017towards}
Huang Y, Bogo F, Lassner C, Kanazawa A, Gehler PV, Romero J, Akhter I, Black MJ (2017) Towards accurate marker-less human shape and pose estimation over time. In: 2017 international conference on 3D vision (3DV), IEEE, pp 421--430

\bibitem[{Huang et~al.(2018)Huang, Kaufmann, Aksan, Black, Hilliges, and Pons-Moll}]{huang2018DIP}
Huang Y, Kaufmann M, Aksan E, Black MJ, Hilliges O, Pons-Moll G (2018) Deep inertial poser: Learning to reconstruct human pose from sparse inertial measurements in real time. ACM Transactions on Graphics (TOG) 37(6):1--15

\bibitem[{Jiang et~al.(2023)Jiang, Chen, Liu, Yu, Yu, and Chen}]{jiang2023motiongpt}
Jiang B, Chen X, Liu W, Yu J, Yu G, Chen T (2023) Motiongpt: Human motion as a foreign language. arXiv preprint arXiv:230614795

\bibitem[{Kalakonda et~al.(2023)Kalakonda, Maheshwari, and Sarvadevabhatla}]{kalakonda2023action}
Kalakonda SS, Maheshwari S, Sarvadevabhatla RK (2023) Action-gpt: Leveraging large-scale language models for improved and generalized action generation. In: 2023 IEEE International Conference on Multimedia and Expo (ICME), IEEE, pp 31--36

\bibitem[{Kanazawa et~al.(2019)Kanazawa, Zhang, Felsen, and Malik}]{kanazawa2019learning}
Kanazawa A, Zhang JY, Felsen P, Malik J (2019) Learning 3d human dynamics from video. In: Proceedings of the IEEE/CVF conference on computer vision and pattern recognition, pp 5614--5623

\bibitem[{Kenton and Toutanova(2019)}]{devlin2018bert}
Kenton JDMWC, Toutanova LK (2019) Bert: Pre-training of deep bidirectional transformers for language understanding. In: Proceedings of naacL-HLT, vol~1, p~2

\bibitem[{Kim et~al.(2023)Kim, Kim, and Choi}]{kim2022flame}
Kim J, Kim J, Choi S (2023) Flame: Free-form language-based motion synthesis \& editing. In: Proceedings of the AAAI Conference on Artificial Intelligence, vol~37, pp 8255--8263

\bibitem[{Kingma and Welling(2013)}]{kingma2013auto}
Kingma DP, Welling M (2013) Auto-encoding variational bayes. arXiv preprint arXiv:13126114

\bibitem[{Kocabas et~al.(2020)Kocabas, Athanasiou, and Black}]{kocabas2020vibe}
Kocabas M, Athanasiou N, Black MJ (2020) Vibe: Video inference for human body pose and shape estimation. In: Proceedings of the IEEE/CVF conference on computer vision and pattern recognition, pp 5253--5263

\bibitem[{Kolotouros et~al.(2019)Kolotouros, Pavlakos, Black, and Daniilidis}]{kolotouros2019learning}
Kolotouros N, Pavlakos G, Black MJ, Daniilidis K (2019) Learning to reconstruct 3d human pose and shape via model-fitting in the loop. In: Proceedings of the IEEE/CVF international conference on computer vision, pp 2252--2261

\bibitem[{Lassner et~al.(2017)Lassner, Romero, Kiefel, Bogo, Black, and Gehler}]{lassner2017unite}
Lassner C, Romero J, Kiefel M, Bogo F, Black MJ, Gehler PV (2017) Unite the people: Closing the loop between 3d and 2d human representations. In: Proceedings of the IEEE conference on computer vision and pattern recognition, pp 6050--6059

\bibitem[{Lee et~al.(2019)Lee, Yang, Liu, Wang, Lu, Yang, and Kautz}]{lee2019dancing}
Lee HY, Yang X, Liu MY, Wang TC, Lu YD, Yang MH, Kautz J (2019) Dancing to music. Advances in neural information processing systems 32

\bibitem[{Li et~al.(2022)Li, Zhao, Zhelun, and Sheng}]{li2022danceformer}
Li B, Zhao Y, Zhelun S, Sheng L (2022) Danceformer: Music conditioned 3d dance generation with parametric motion transformer. In: Proceedings of the AAAI Conference on Artificial Intelligence, vol~36, pp 1272--1279

\bibitem[{Li et~al.(2021)Li, Yang, Ross, and Kanazawa}]{li2021ai}
Li R, Yang S, Ross DA, Kanazawa A (2021) Ai choreographer: Music conditioned 3d dance generation with aist++. In: Proceedings of the IEEE/CVF International Conference on Computer Vision, pp 13401--13412

\bibitem[{Liang et~al.(2023)Liang, He, Zhao, Li, Wang, Yu, and Xu}]{liang2022hybridcap}
Liang H, He Y, Zhao C, Li M, Wang J, Yu J, Xu L (2023) Hybridcap: Inertia-aid monocular capture of challenging human motions. In: Proceedings of the AAAI Conference on Artificial Intelligence, vol~37, pp 1539--1548

\bibitem[{Liu et~al.(2019)Liu, Shahroudy, Perez, Wang, Duan, and Kot}]{liu2019ntu}
Liu J, Shahroudy A, Perez M, Wang G, Duan LY, Kot AC (2019) Ntu rgb+ d 120: A large-scale benchmark for 3d human activity understanding. IEEE transactions on pattern analysis and machine intelligence 42(10):2684--2701

\bibitem[{Liu et~al.(2013)Liu, Gall, Stoll, Dai, Seidel, and Theobalt}]{liu2013markerless}
Liu Y, Gall J, Stoll C, Dai Q, Seidel HP, Theobalt C (2013) Markerless motion capture of multiple characters using multiview image segmentation. IEEE transactions on pattern analysis and machine intelligence 35(11):2720--2735

\bibitem[{Loper et~al.(2015)Loper, Mahmood, Romero, Pons-Moll, and Black}]{loper2015smpl}
Loper M, Mahmood N, Romero J, Pons-Moll G, Black MJ (2015) Smpl: A skinned multi-person linear model. ACM transactions on graphics (TOG) 34(6):1--16

\bibitem[{Loshchilov and Hutter(2018)}]{loshchilov2017decoupled}
Loshchilov I, Hutter F (2018) Decoupled weight decay regularization. In: International Conference on Learning Representations

\bibitem[{Lucas et~al.(2022)Lucas, Baradel, Weinzaepfel, and Rogez}]{lucas2022posegpt}
Lucas T, Baradel F, Weinzaepfel P, Rogez G (2022) Posegpt: Quantization-based 3d human motion generation and forecasting. In: European Conference on Computer Vision, Springer, pp 417--435

\bibitem[{Malleson et~al.(2017)Malleson, Gilbert, Trumble, Collomosse, Hilton, and Volino}]{malleson2017real}
Malleson C, Gilbert A, Trumble M, Collomosse J, Hilton A, Volino M (2017) Real-time full-body motion capture from video and imus. In: 2017 international conference on 3D vision (3DV), IEEE, pp 449--457

\bibitem[{Malleson et~al.(2019)Malleson, Collomosse, and Hilton}]{malleson2019real}
Malleson C, Collomosse J, Hilton A (2019) Real-time multi-person motion capture from multi-view video and imus. International Journal of Computer Vision pp 1--18

\bibitem[{Movella(2022)}]{XSENS}
Movella (2022) Movella xsens products. \url{https://www.movella.com/products/xsens}, accessed: 2023-03-26

\bibitem[{Ng et~al.(2020)Ng, Xiang, Joo, and Grauman}]{ng2020you2me}
Ng E, Xiang D, Joo H, Grauman K (2020) You2me: Inferring body pose in egocentric video via first and second person interactions. In: Proceedings of the IEEE/CVF Conference on Computer Vision and Pattern Recognition, pp 9890--9900

\bibitem[{Nichol and Dhariwal(2021)}]{nichol2021improved}
Nichol AQ, Dhariwal P (2021) Improved denoising diffusion probabilistic models. In: International Conference on Machine Learning, PMLR, pp 8162--8171

\bibitem[{OpenAI(2023)}]{openai2023gpt4}
OpenAI (2023) Gpt-4 technical report. \eprint{2303.08774}

\bibitem[{Osman et~al.(2020)Osman, Bolkart, and Black}]{osman2020star}
Osman AA, Bolkart T, Black MJ (2020) Star: Sparse trained articulated human body regressor. In: Computer Vision--ECCV 2020: 16th European Conference, Glasgow, UK, August 23--28, 2020, Proceedings, Part VI 16, Springer, pp 598--613

\bibitem[{Pavlakos et~al.(2017)Pavlakos, Zhou, Derpanis, and Daniilidis}]{Pavlakos17}
Pavlakos G, Zhou X, Derpanis KG, Daniilidis K (2017) Harvesting multiple views for marker-less 3d human pose annotations. In: Computer Vision and Pattern Recognition (CVPR)

\bibitem[{Pavlakos et~al.(2019)Pavlakos, Choutas, Ghorbani, Bolkart, Osman, Tzionas, and Black}]{pavlakos2019expressive}
Pavlakos G, Choutas V, Ghorbani N, Bolkart T, Osman AA, Tzionas D, Black MJ (2019) Expressive body capture: 3d hands, face, and body from a single image. In: Proceedings of the IEEE/CVF conference on computer vision and pattern recognition, pp 10975--10985

\bibitem[{Peng et~al.(2021)Peng, Ma, Abbeel, Levine, and Kanazawa}]{peng2021amp}
Peng XB, Ma Z, Abbeel P, Levine S, Kanazawa A (2021) Amp: Adversarial motion priors for stylized physics-based character control. ACM Transactions on Graphics (TOG) 40(4):1--20

\bibitem[{Petrovich et~al.(2021)Petrovich, Black, and Varol}]{petrovich2021action}
Petrovich M, Black MJ, Varol G (2021) Action-conditioned 3d human motion synthesis with transformer vae. In: Proceedings of the IEEE/CVF International Conference on Computer Vision, pp 10985--10995

\bibitem[{Petrovich et~al.(2022)Petrovich, Black, and Varol}]{petrovich2022temos}
Petrovich M, Black MJ, Varol G (2022) Temos: Generating diverse human motions from textual descriptions. In: Computer Vision--ECCV 2022: 17th European Conference, Tel Aviv, Israel, October 23--27, 2022, Proceedings, Part XXII, Springer, pp 480--497

\bibitem[{Plappert et~al.(2016)Plappert, Mandery, and Asfour}]{Plappert2016kit}
Plappert M, Mandery C, Asfour T (2016) The kit motion-language dataset. Big data 4(4):236--252

\bibitem[{Punnakkal et~al.(2021)Punnakkal, Chandrasekaran, Athanasiou, Quiros-Ramirez, and Black}]{punnakkal2021babel}
Punnakkal AR, Chandrasekaran A, Athanasiou N, Quiros-Ramirez A, Black MJ (2021) Babel: bodies, action and behavior with english labels. In: Proceedings of the IEEE/CVF Conference on Computer Vision and Pattern Recognition, pp 722--731

\bibitem[{Radford et~al.(2021)Radford, Kim, Hallacy, Ramesh, Goh, Agarwal, Sastry, Askell, Mishkin, Clark et~al.}]{radford2021learning}
Radford A, Kim JW, Hallacy C, Ramesh A, Goh G, Agarwal S, Sastry G, Askell A, Mishkin P, Clark J, et~al. (2021) Learning transferable visual models from natural language supervision. In: International conference on machine learning, PMLR, pp 8748--8763

\bibitem[{Rempe et~al.(2021)Rempe, Birdal, Hertzmann, Yang, Sridhar, and Guibas}]{rempe2021humor}
Rempe D, Birdal T, Hertzmann A, Yang J, Sridhar S, Guibas LJ (2021) Humor: 3d human motion model for robust pose estimation. In: Proceedings of the IEEE/CVF international conference on computer vision, pp 11488--11499

\bibitem[{Ren et~al.(2023)Ren, Zhao, He, Cong, Liang, Yu, Xu, and Ma}]{zhao2022lidar}
Ren Y, Zhao C, He Y, Cong P, Liang H, Yu J, Xu L, Ma Y (2023) Lidar-aid inertial poser: Large-scale human motion capture by sparse inertial and lidar sensors. IEEE Transactions on Visualization and Computer Graphics 29(5):2337--2347

\bibitem[{Rezende and Mohamed(2015)}]{rezende2015variational}
Rezende D, Mohamed S (2015) Variational inference with normalizing flows. In: International conference on machine learning, PMLR, pp 1530--1538

\bibitem[{Robertini et~al.(2016)Robertini, Casas, Rhodin, Seidel, and Theobalt}]{robertini2016model}
Robertini N, Casas D, Rhodin H, Seidel HP, Theobalt C (2016) Model-based outdoor performance capture. In: 2016 Fourth International Conference on 3D Vision (3DV), IEEE, pp 166--175

\bibitem[{Shafir et~al.(2023)Shafir, Tevet, Kapon, and Bermano}]{shafir2023human}
Shafir Y, Tevet G, Kapon R, Bermano AH (2023) Human motion diffusion as a generative prior. arXiv preprint arXiv:230301418

\bibitem[{Simon et~al.(2017)Simon, Joo, Matthews, and Sheikh}]{Simon17}
Simon T, Joo H, Matthews I, Sheikh Y (2017) Hand keypoint detection in single images using multiview bootstrapping. In: Computer Vision and Pattern Recognition (CVPR)

\bibitem[{Song et~al.(2020{\natexlab{a}})Song, Meng, and Ermon}]{song2020denoising}
Song J, Meng C, Ermon S (2020{\natexlab{a}}) Denoising diffusion implicit models. In: International Conference on Learning Representations

\bibitem[{Song et~al.(2020{\natexlab{b}})Song, Sohl-Dickstein, Kingma, Kumar, Ermon, and Poole}]{song2020score}
Song Y, Sohl-Dickstein J, Kingma DP, Kumar A, Ermon S, Poole B (2020{\natexlab{b}}) Score-based generative modeling through stochastic differential equations. In: International Conference on Learning Representations

\bibitem[{Starke et~al.(2019)Starke, Zhang, Komura, and Saito}]{starke2019neural}
Starke S, Zhang H, Komura T, Saito J (2019) Neural state machine for character-scene interactions. ACM Trans Graph 38(6):209--1

\bibitem[{Starke et~al.(2022)Starke, Mason, and Komura}]{starke2022deepphase}
Starke S, Mason I, Komura T (2022) Deepphase: Periodic autoencoders for learning motion phase manifolds. ACM Transactions on Graphics (TOG) 41(4):1--13

\bibitem[{Stoll et~al.(2011)Stoll, Hasler, Gall, Seidel, and Theobalt}]{StollHGST2011}
Stoll C, Hasler N, Gall J, Seidel HP, Theobalt C (2011) Fast articulated motion tracking using a sums of {Gaussians} body model. In: International Conference on Computer Vision (ICCV)

\bibitem[{Tanaka and Fujiwara(2023)}]{tanaka2023role}
Tanaka M, Fujiwara K (2023) Role-aware interaction generation from textual description. In: Proceedings of the IEEE/CVF International Conference on Computer Vision, pp 15999--16009

\bibitem[{Tevet et~al.(2022{\natexlab{a}})Tevet, Gordon, Hertz, Bermano, and Cohen-Or}]{tevet2022motionclip}
Tevet G, Gordon B, Hertz A, Bermano AH, Cohen-Or D (2022{\natexlab{a}}) Motionclip: Exposing human motion generation to clip space. In: Computer Vision--ECCV 2022: 17th European Conference, Tel Aviv, Israel, October 23--27, 2022, Proceedings, Part XXII, Springer, pp 358--374

\bibitem[{Tevet et~al.(2022{\natexlab{b}})Tevet, Raab, Gordon, Shafir, Cohen-Or, and Bermano}]{tevet2022human}
Tevet G, Raab S, Gordon B, Shafir Y, Cohen-Or D, Bermano AH (2022{\natexlab{b}}) Human motion diffusion model. In: International Conference on Learning Representations

\bibitem[{Theobalt et~al.(2010)Theobalt, de~Aguiar, Stoll, Seidel, and Thrun}]{theobalt2010performance}
Theobalt C, de~Aguiar E, Stoll C, Seidel HP, Thrun S (2010) Performance capture from multi-view video. In: Image and Geometry Processing for 3-D Cinematography, Springer, pp 127--149

\bibitem[{Touvron et~al.(2023)Touvron, Lavril, Izacard, Martinet, Lachaux, Lacroix, Rozi{\`e}re, Goyal, Hambro, Azhar et~al.}]{touvron2023llama}
Touvron H, Lavril T, Izacard G, Martinet X, Lachaux MA, Lacroix T, Rozi{\`e}re B, Goyal N, Hambro E, Azhar F, et~al. (2023) Llama: Open and efficient foundation language models. arXiv preprint arXiv:230213971

\bibitem[{Vicon(2019)}]{VICON}
Vicon (2019) {Vicon Motion Systems}. \url{https://www.vicon.com/}

\bibitem[{Vlasic et~al.(2007)Vlasic, Adelsberger, Vannucci, Barnwell, Gross, Matusik, and Popovi{\'c}}]{vlasic2007practical}
Vlasic D, Adelsberger R, Vannucci G, Barnwell J, Gross M, Matusik W, Popovi{\'c} J (2007) Practical motion capture in everyday surroundings. ACM transactions on graphics (TOG) 26(3):35--es

\bibitem[{Von~Marcard et~al.(2017)Von~Marcard, Rosenhahn, Black, and Pons-Moll}]{von2017SIP}
Von~Marcard T, Rosenhahn B, Black MJ, Pons-Moll G (2017) Sparse inertial poser: Automatic 3d human pose estimation from sparse imus. In: Computer Graphics Forum, Wiley Online Library, vol~36, pp 349--360

\bibitem[{Von~Marcard et~al.(2018)Von~Marcard, Henschel, Black, Rosenhahn, and Pons-Moll}]{von2018recovering}
Von~Marcard T, Henschel R, Black MJ, Rosenhahn B, Pons-Moll G (2018) Recovering accurate 3d human pose in the wild using imus and a moving camera. In: Proceedings of the European conference on computer vision (ECCV), pp 601--617

\bibitem[{Wang et~al.(2021)Wang, Yan, Dai, and Lin}]{wang2021scene}
Wang J, Yan S, Dai B, Lin D (2021) Scene-aware generative network for human motion synthesis. In: Proceedings of the IEEE/CVF Conference on Computer Vision and Pattern Recognition, pp 12206--12215

\bibitem[{Wang et~al.(2022)Wang, Chen, Liu, Zhu, Liang, and Huang}]{wang2022humanise}
Wang Z, Chen Y, Liu T, Zhu Y, Liang W, Huang S (2022) Humanise: Language-conditioned human motion generation in 3d scenes. Advances in Neural Information Processing Systems 35:14959--14971

\bibitem[{Xu et~al.(2018{\natexlab{a}})Xu, Liu, Cheng, Guo, Zhou, Dai, and Fang}]{FlyCap}
Xu L, Liu Y, Cheng W, Guo K, Zhou G, Dai Q, Fang L (2018{\natexlab{a}}) Flycap: Markerless motion capture using multiple autonomous flying cameras. IEEE Transactions on Visualization and Computer Graphics 24(8):2284--2297

\bibitem[{Xu et~al.(2020)Xu, Xu, Golyanik, Habermann, Fang, and Theobalt}]{EventCap_CVPR2020}
Xu L, Xu W, Golyanik V, Habermann M, Fang L, Theobalt C (2020) Eventcap: Monocular 3d capture of high-speed human motions using an event camera. In: Proceedings of the IEEE/CVF Conference on Computer Vision and Pattern Recognition, pp 4968--4978

\bibitem[{Xu et~al.(2023)Xu, Song, Wang, Su, Fang, Ding, Gan, Yan, Jin, Yang et~al.}]{song2022actformer}
Xu L, Song Z, Wang D, Su J, Fang Z, Ding C, Gan W, Yan Y, Jin X, Yang X, et~al. (2023) Actformer: A gan-based transformer towards general action-conditioned 3d human motion generation. In: Proceedings of the IEEE/CVF International Conference on Computer Vision, pp 2228--2238

\bibitem[{Xu et~al.(2018{\natexlab{b}})Xu, Chatterjee, Zollh\"{o}fer, Rhodin, Mehta, Seidel, and Theobalt}]{MonoPerfCap}
Xu W, Chatterjee A, Zollh\"{o}fer M, Rhodin H, Mehta D, Seidel HP, Theobalt C (2018{\natexlab{b}}) Monoperfcap: Human performance capture from monocular video. ACM Transactions on Graphics (TOG) 37(2):27:1--27:15

\bibitem[{Yi et~al.(2021)Yi, Zhou, and Xu}]{TransPose2021}
Yi X, Zhou Y, Xu F (2021) Transpose: Real-time 3d human translation and pose estimation with six inertial sensors. ACM Transactions on Graphics (TOG) 40(4):1--13

\bibitem[{Yi et~al.(2022)Yi, Zhou, Habermann, Shimada, Golyanik, Theobalt, and Xu}]{PIPCVPR2022}
Yi X, Zhou Y, Habermann M, Shimada S, Golyanik V, Theobalt C, Xu F (2022) Physical inertial poser (pip): Physics-aware real-time human motion tracking from sparse inertial sensors. In: IEEE/CVF Conference on Computer Vision and Pattern Recognition (CVPR)

\bibitem[{You et~al.(2020)You, Leskovec, He, and Xie}]{you2020graph}
You J, Leskovec J, He K, Xie S (2020) Graph structure of neural networks. In: International Conference on Machine Learning, PMLR, pp 10881--10891

\bibitem[{Yuan et~al.(2023)Yuan, Song, Iqbal, Vahdat, and Kautz}]{yuan2022physdiff}
Yuan Y, Song J, Iqbal U, Vahdat A, Kautz J (2023) Physdiff: Physics-guided human motion diffusion model. In: Proceedings of the IEEE/CVF International Conference on Computer Vision, pp 16010--16021

\bibitem[{Z-cam(2022)}]{Z-CAM}
Z-cam (2022) {Z CAM Cinema Camera}. \url{https://www.z-cam.com}, accessed :2023-03-26

\bibitem[{Zanfir et~al.(2021)Zanfir, Bazavan, Zanfir, Freeman, Sukthankar, and Sminchisescu}]{zanfir2020neural}
Zanfir A, Bazavan EG, Zanfir M, Freeman WT, Sukthankar R, Sminchisescu C (2021) Neural descent for visual 3d human pose and shape. In: Proceedings of the IEEE/CVF Conference on Computer Vision and Pattern Recognition, pp 14484--14493

\bibitem[{Zhang et~al.(2023{\natexlab{a}})Zhang, Zhang, Cun, Zhang, Zhao, Lu, Shen, and Shan}]{zhang2023t2m}
Zhang J, Zhang Y, Cun X, Zhang Y, Zhao H, Lu H, Shen X, Shan Y (2023{\natexlab{a}}) Generating human motion from textual descriptions with discrete representations. In: Proceedings of the IEEE/CVF Conference on Computer Vision and Pattern Recognition, pp 14730--14740

\bibitem[{Zhang et~al.(2022)Zhang, Cai, Pan, Hong, Guo, Yang, and Liu}]{zhang2022motiondiffuse}
Zhang M, Cai Z, Pan L, Hong F, Guo X, Yang L, Liu Z (2022) Motiondiffuse: Text-driven human motion generation with diffusion model. arXiv preprint arXiv:220815001

\bibitem[{Zhang et~al.(2023{\natexlab{b}})Zhang, Huang, Liu, Tang, Lu, Chen, Bai, Chu, Yu, and Ouyang}]{zhang2023motiongpt}
Zhang Y, Huang D, Liu B, Tang S, Lu Y, Chen L, Bai L, Chu Q, Yu N, Ouyang W (2023{\natexlab{b}}) Motiongpt: Finetuned llms are general-purpose motion generators. arXiv preprint arXiv:230610900

\bibitem[{Zheng et~al.(2018)Zheng, Yu, Li, Guo, Dai, Fang, and Liu}]{Zheng2018HybridFusion}
Zheng Z, Yu T, Li H, Guo K, Dai Q, Fang L, Liu Y (2018) Hybridfusion: Real-time performance capture using a single depth sensor and sparse imus. In: Proceedings of the European Conference on Computer Vision (ECCV), pp 384--400

\end{thebibliography}
}






\end{document}